\documentclass[10pt,twocolumn,letterpaper]{article}

\usepackage{cvpr}              

\usepackage{graphicx}
\usepackage{amsmath}
\usepackage{amssymb}
\usepackage{booktabs}
\usepackage{float}
\usepackage{multirow}
\usepackage{booktabs}

\usepackage{placeins}
\usepackage{paralist}

\usepackage[pagebackref,breaklinks,colorlinks]{hyperref}

\usepackage[capitalize]{cleveref}
\crefname{section}{Sec.}{Secs.}
\Crefname{section}{Section}{Sections}
\Crefname{table}{Table}{Tables}
\crefname{table}{Tab.}{Tabs.}

\def\cvprPaperID{*****} 
\def\confName{CVPR}
\def\confYear{2022}

\begin{document}
\title{Self-supervised Learning for Sonar Image Classification}

\author{
\parbox{\linewidth}{
    Alan Preciado-Grijalva $^{1,2}$, Bilal Wehbe$^1$, Miguel Bande Firvida$^1$, Matias Valdenegro-Toro$^{1,3}$\\
}\\
\parbox{\linewidth}{
    $^1$ German Research Center for Artificial Intelligence, 28359 Bremen, Germany.\\
    $^2$ Bonn-Rhein-Sieg University of Applied Sciences, 53757 Sankt Augustin, Germany.\\
    $^3$ Department of AI, University of Groningen, 9747AG Groningen, The Netherlands.\\
}\\
\parbox{\linewidth}{
    {\tt\small agrija9@gmail.com, [bilal.wehbe, miguel.bande\_firvida]@dfki.de, m.a.valdenegro.toro@rug.nl}
}
}
\maketitle


\begin{abstract}
Self-supervised learning has proved to be a powerful approach to learn image representations without the need of large labeled datasets. For underwater robotics, it is of great interest to design computer vision algorithms to improve perception capabilities such as sonar image classification. Due to the confidential nature of sonar imaging and the difficulty to interpret sonar images, it is challenging to create public large labeled  sonar datasets to train supervised learning algorithms. In this work, we investigate the potential of three self-supervised learning methods (RotNet, Denoising Autoencoders, and Jigsaw) to learn high-quality sonar image representation without the need of human labels. We present pre-training and transfer learning results on real-life sonar image datasets. Our results indicate that self-supervised pre-training yields classification performance comparable to supervised pre-training in a few-shot transfer learning setup across all three methods. Code and self-supervised pre-trained models are be available at \href{https://github.com/agrija9/ssl-sonar-images}{agrija9/ssl-sonar-images}.

\end{abstract}

\section{Introduction}

\label{sec:intro}

Machine perception in autonomous underwater systems is considered an exceptionally challenging task due to the unpredictability of marine environments. Factors such as poor lighting conditions, sediments, and turbidity impact the visual-based sensing of underwater systems and can result in failure of critical localization and exploration missions. Acoustic-based imaging provides an alternative sensing modality that is unimpeded by visibility conditions and could operate in turbid waters or complete darkness. The main challenges faced however with acoustic sensors such as imaging sonars are low signal-to-noise ratios and other disturbances such as acoustic shadows, multipath interference \cite{saucan2015model} and crosstalk noise \cite{sung2018crosstalk}.

In recent years, deep neural network (DNN) architectures have become more popular in underwater applications like image enhancement \cite{kim2017denoising,sung2018crosstalk}, sonar object classification \cite{valdenegro2016object,Matias_Valdenegro_IV,wang2019underwater}, and sonar-camera image translation \cite{terayama2019integration,jang2019cnn}. For underwater sonar data specifically, the progress of deep learning research has been hindered due to the lack of publicly available data, which could be contributed to the costly operations needed to collect sonar data as well as confidentiality issues regarding military applications. To compensate for small datasets and lack of annotated data, self-supervised learning (SSL) \cite{Jing_2020, jaiswal2021survey, liu2021self} has become an effective technique that attempts to learn data representations by using data itself as a supervision signal.

		\begin{figure}
		\centering
		\includegraphics[scale=0.5]{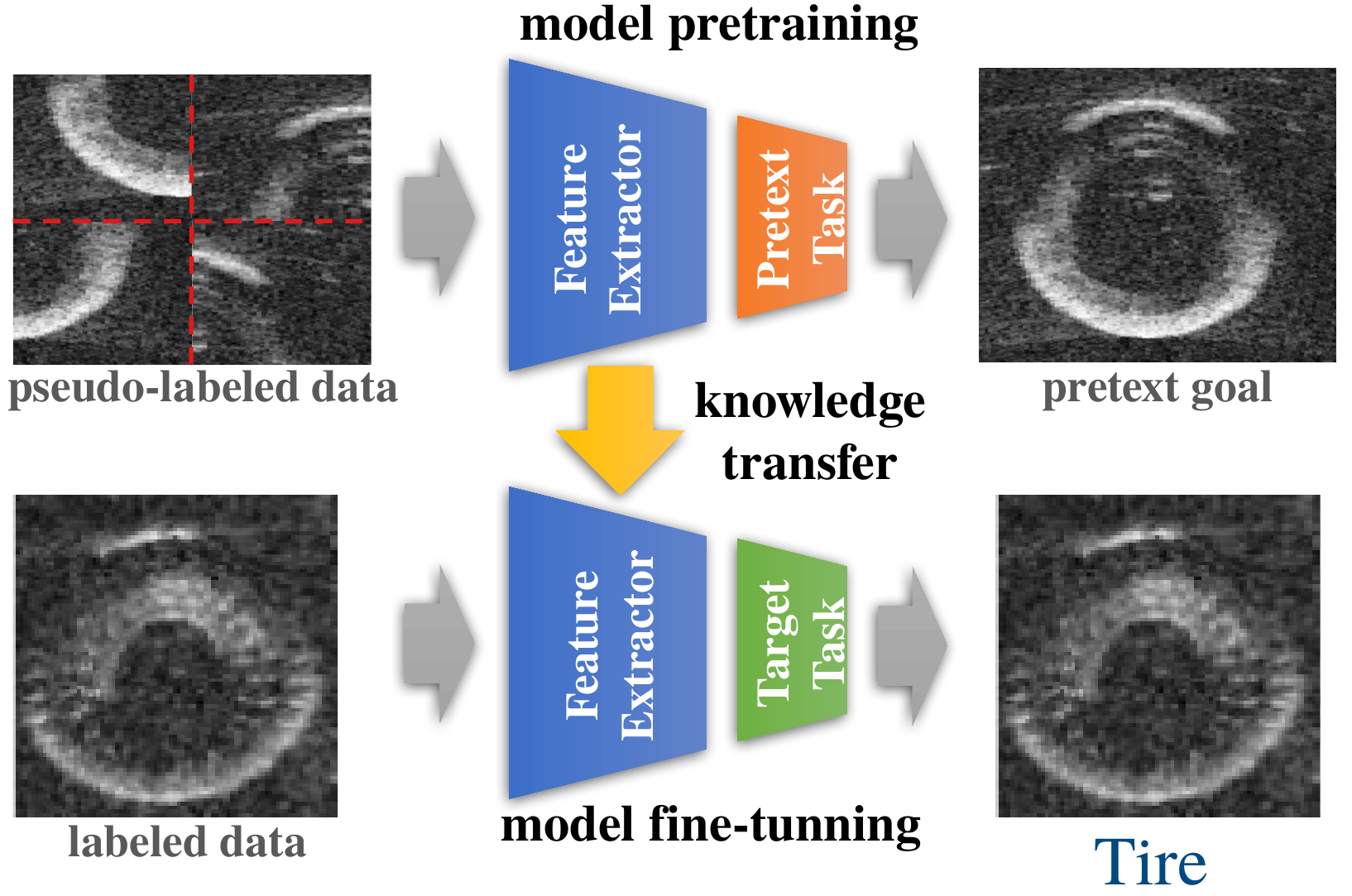}
		  \caption{We propose to expand the capabilities of self-supervised learning to the sonar image domain with applications to underwater robotics. Our conceptual idea of self-supervision on grayscale sonar images follows successful established approaches where we pretrain models on large unlabeled sonar datasets and evaluate them using transfer learning on smaller labeled sonar datasets.}
		  \label{fig:ssl_concept_sonar_images}
		\end{figure}

The main contribution of this work is the evaluation of relevant SSL algorithms for sonar image data with applications to object classification. In \cref{fig:ssl_concept_sonar_images} we illustrate our proposed conceptual approach. We have studied three SSL algorithms (RotNet \cite{Gidaris_I}, Denoising Autoencoders \cite{Pascal_I}, and Jigsaw \cite{Noroozi_I}) and compared them to their supervised learning (SL) counterpart. The SSL algorithms were trained on an experimental real-life sonar dataset, where the learned representation quality was evaluated using a low-shot transfer learning setup on another real-life test sonar dataset. Our findings indicate that all SSL models can reach a similar performance to that of their supervised counterparts. This holds even when using a completely unlabeled on-the-wild sonar dataset. We also note that unlike color image datasets, SSL does not by itself outperform supervised learning. 

These results indicate that SSL has the potential to replace the need for labeled sonar data without compromising task performance and reducing the time and costs of data labeling.

\section{Related Work}
\label{sec:related_work}

Self-supervised learning (SSL) is a learning paradigm that presents itself as a more scalable approach to pretrain models for transfer learning with less human labeled data \cite{Mathilde_Caron, Ting_Chen, Jean_Grill, Olivier_Henaff}. SSL can be understood as a two-step approach:
\begin{inparaenum}[a)]
  \item learning data representation from solving a proxy or pretext task using automatically generated pseudo-labels from raw-unlabeled data, followed by
  \item fine-tuning of the learned features on the actual downstream task, i.e. task of interest such as classification, segmentation or detection tasks, with few manually labeled data by using transfer learning.
\end{inparaenum}

SSL has been mainly used in color images demonstrating not only the capability to reach same performance than pure supervised learning methods with much less labeled data for many tasks, but even surpassing their performance in some cases \cite{Jing_2020}. Pretext tasks that have shown promising results in learning strong latent representations in color images are
\begin{inparaenum}[a)]
  \item generation-based methods, such as image colorization \cite{Zhang_I, Zhang_III, Larsson_I}, super-resolution \cite{Ledig_I}, image denoising \cite{Pascal_I}, in-painting \cite{Pathak_I, Satoshi_I} and video prediction \cite{Srivastava_2015},
  \item spatial context methods, such as solving the jigsaw puzzle \cite{Noroozi_I, Taleb_2021} and recognizing rotations \cite{Gidaris_I}, and
  \item temporal context methods, such as recognizing the order of the frame sequence \cite{Misra_2016, Lee_2017}.
\end{inparaenum}
A detailed description of the chosen pretext tasks widely used for color images that we believe can be applied for sonar images are given in section \ref{sec:sonar_classification}.

Although self-supervised learning has been demonstrated to improve performance in color images for many downstream tasks, it has not yet been applied to sonar images to the best of our knowledge. However, a few studies on supervised pre-training for transfer learning have been carried out showing great potential for the use of SSL with sonar images \cite{Matias_Valdenegro_I, valdenegro2016object, Matias_Valdenegro_IV}. For example, \cite{Matias_Valdenegro_I} made a study on the effect of the training set size by using transfer learning in sonar image for object recognition, some of the proposed architectures required only 50 samples per class to achieve 90\% accuracy indicating that deep convolutional neural network (deep CNN) models can generalize well to other data distributions even with few samples per class.

\section{Sonar Datasets}
\label{sec:sonar_datasets}

One of the most important things to build robust and reliable deep learning vision models is to test them across different datasets in order to see how they perform against varying data distributions. In this section, we describe the sonar datasets we used to train and evaluate our self-supervised learning algorithms. In particular, we used the \textit{Marine Debris Watertank} dataset in the pre-training phase and the \textit{Marine Debris Turntable} dataset to evaluate the quality of the learned features during pre-training. Both datasets were introduced in \cite{Matias_Valdenegro_IV}. 

\begin{figure}
		\centering
		\includegraphics[scale=0.42]{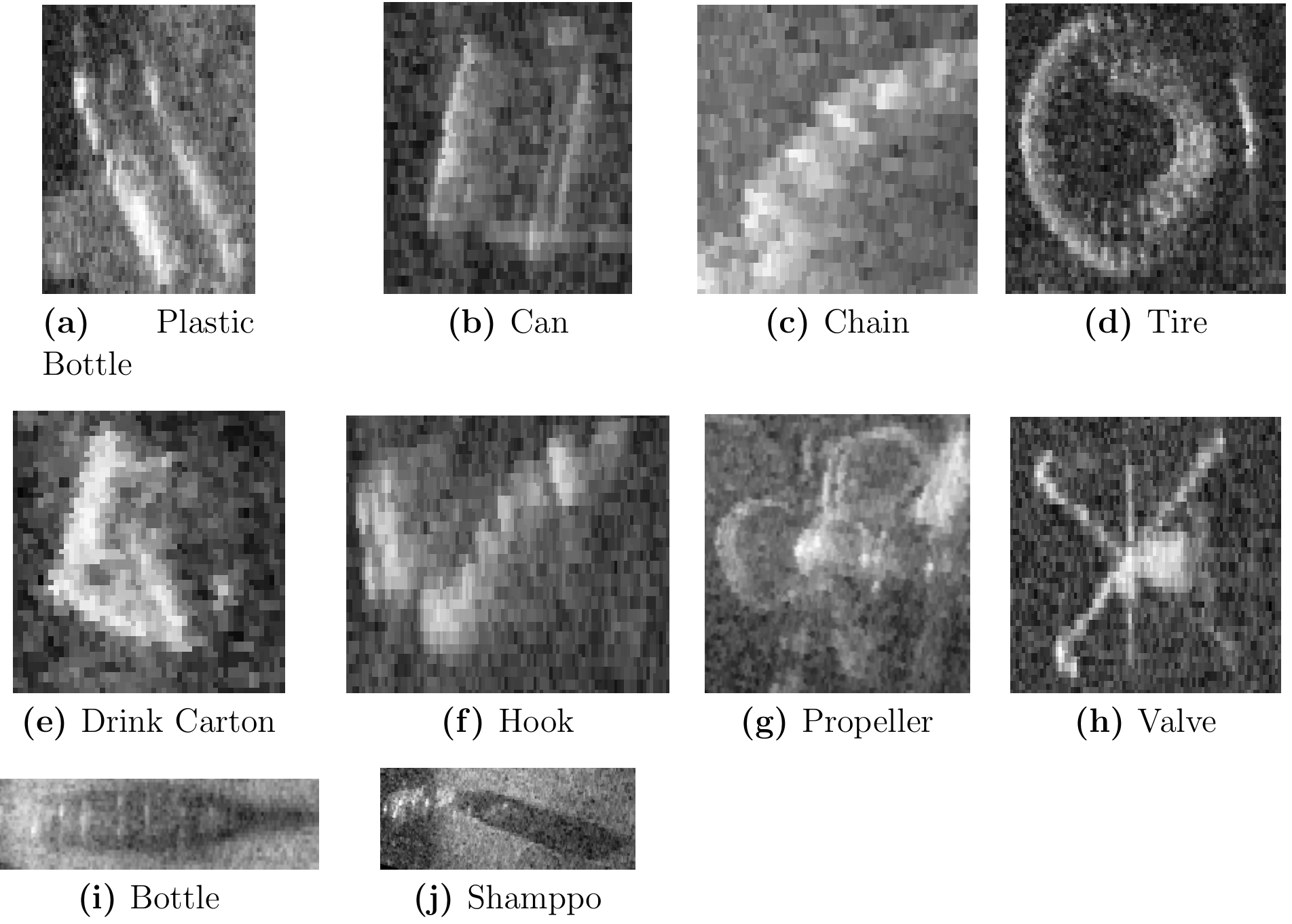}
		  \captionof{figure}{Samples of sonar images from the Watertank dataset.}
		  \label{fig:watertank_data}
		\end{figure}

\textbf{Marine Debris Watertank}. This dataset contains a total of 2627 forward-looking sonar (FLS) images grouped across 11 classes. This dataset was collected with an ARIS Explorer 3000 FLS at a frequency of 3.0 MHz. The dataset was split into three sets: 70\% for training, 15\% for validation and 15\% for testing.  We used this dataset exclusively to pretrain the three self-supervised models proposed in this paper.
\cref{fig:watertank_data} shows objects sampled from this dataset.

\begin{figure}
		\centering
		  \includegraphics[scale=0.45]{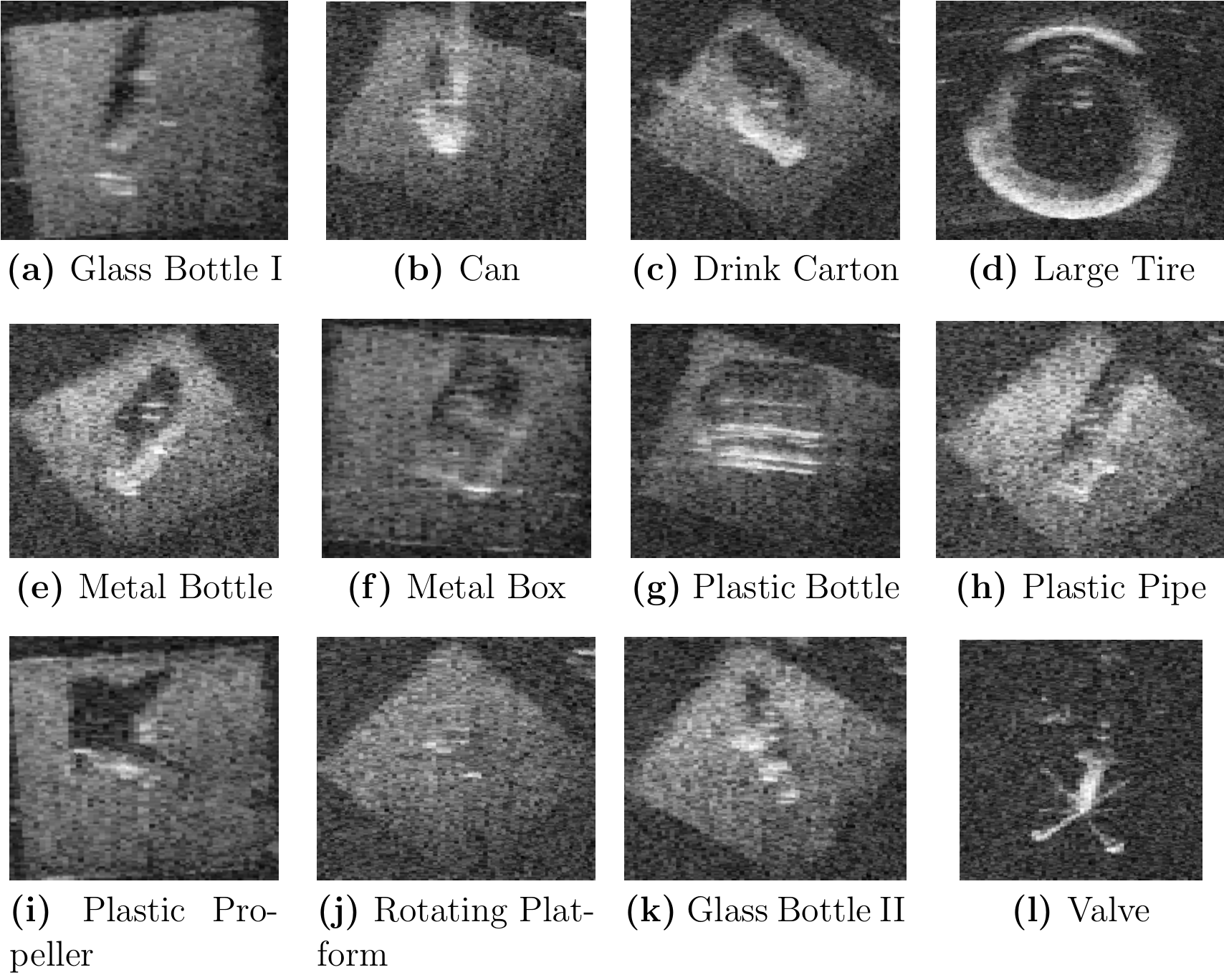}
		  \captionof{figure}{Samples of sonar images from the Turntable dataset.}
		  \label{fig:turntable_data}
		\end{figure}

\textbf{Marine Debris Turntable}. This dataset contains a total of 2471 sonar images grouped across 12 classes. This dataset was also collected using an ARIS Explorer 3000 FLS at the highest frequency. Each object in this dataset was placed underwater on a rotating table so that the images for each object were captured at different angles along the z-axis. As mentioned in \cite{Matias_Valdenegro_IV}, generating multiple views from objects can help learn better image features since sonar image properties changes with the view angle (e.g. reflections, pose, and sensor noise). There is an intersection between both datasets since they have some objects in common (approximately 50\% of the objects). This means that they are not completely independent. \cref{fig:turntable_data} shows samples of the objects on top of the rotating platform.

\section{Sonar Image Classification}
\label{sec:sonar_classification}

In this section, we present pre-training experiments of three self-supervised models: RotNet \cite{Gidaris_I}, Denoising Autoencoders (DAEs) \cite{Pascal_I} and Jigsaw Puzzle \cite{Noroozi_I}. Furthermore, we evaluate their generalization capacities in a few-shot transfer learning setup for sonar image classification. We used the Watertank dataset for pre-training and the Turntable dataset for transfer learning. The research question we address here is whether self-supervised pre-training provides high-quality image features that can compete (or be better) when compared to image features obtained via supervised pre-training.  We have chosen these methods due to their success in applications such as image rotations \cite{Zeyu_Feng, Dan_Hendrycks}, image denoising \cite{Meng_Li, Vladimiros_Sterzentsenko, Yaochen_Xie}, and jigsaw puzzle solving \cite{Priya_goyal, Ru_Li, Ishan_Misra}.

\subsection{RotNet - Learning Sonar Image Representations by Predicting Rotations}
RotNet \cite{Gidaris_I} learns representations by predicting rotation angles applied to input images as opposed to predicting actual object labels. As discussed in \cite{Gidaris_I}, replacing human labels with synthetically generated rotation labels can be an alternative to learn high-quality representations without the need for costly and time-consuming human annotations. Refer to \cref{sec:rotnet_model_selection} for a detailed description of the baseline models we implemented to train RotNet. 

		\begin{figure}
		\centering
		\includegraphics[scale=0.342]{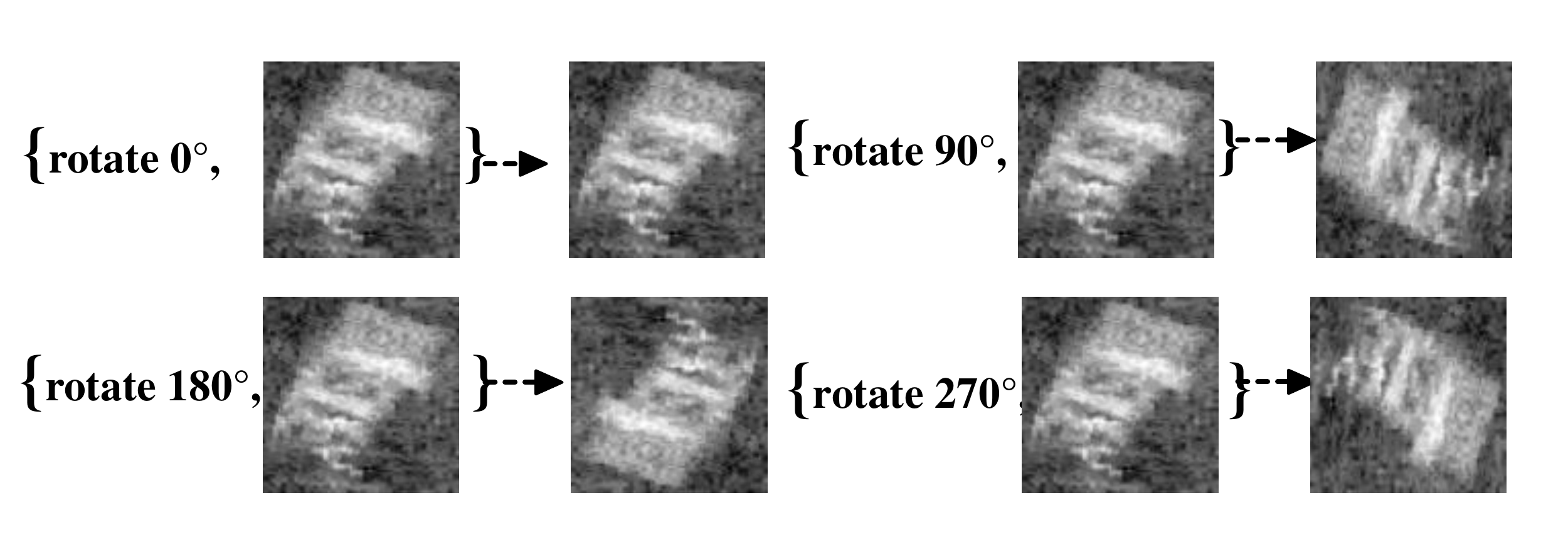}\hfill
		\caption{Applying rotations to sonar images to generate various synthetic perspectives from the same object.}
		\label{fig:rotnet_rotated_samples}
		\end{figure}

\subsubsection{Pre-processing and Hyper-parameter Tuning}

We followed the training procedure from \cite{Gidaris_I} by applying a total of four rotation angles $\{0^o, 90^o, 180^o, 270^o\}$ to the sonar images from the Watertank dataset, see \cref{fig:rotnet_rotated_samples}. Each angle has an integer label associated to it, hence, each image sample has the form $\chi_{rotnet} = $ (image$\_0^o$, 1), (image$\_90^o$, 2), (image$\_180^o$, 3), (image$\_270^o$, 4).


Similar to the experimental procedure from \cite{Matias_Valdenegro_IV}, we tuned a hyper-parameter \textit{w} that corresponds to the number of filters or neurons for each one of the baseline models described previously. These models were originally designed and optimized to perform on standard color images, so it is necessary to fine-tune them for grayscale sonar images. In our case, each architecture has been modified into its shallowest variation, reducing the width parameter \textit{w} in each case. We find that 128 filters as a maximum threshold across all architectures achieve good task performance (as opposed to the 1024 original filters). We found, in agreement with \cite{Matias_Valdenegro_IV}, that the best performing \textit{w} parameters are the ones reported in \cref{table:rotnet_hyperparams_w}.

\begin{table}[]
\scriptsize
\centering
\begin{tabular}{lll}
\toprule
\textbf{Architecture} & \textbf{Selected width w} & \textbf{Use of w in architecture}                                                                                                                                      \\ 
\midrule
ResNet20              & 32                        & \begin{tabular}[c]{@{}l@{}}w is set as starting number of filters,\\ it duplicates itself after each\\ residual stack (w = 2w).\end{tabular}                           \\ \hline
MobileNet             & 32                        & \begin{tabular}[c]{@{}l@{}}Base filters are set to w, increasing\\ first by a factor of four (w = 4w),\\ then duplicating themselves.\end{tabular}                     \\ \hline
DenseNet121           & 16                        & \begin{tabular}[c]{@{}l@{}}w is used as the number of filters\\ for each convolutional layer\\ inside each dense block\end{tabular}                                    \\ \hline
SqueezeNet            & 32                        & \begin{tabular}[c]{@{}l@{}}Base squeeze filters are set to w,\\ and expansion filters are set to 2w,\\ duplicating itself after\\ each fire module stack.\end{tabular} \\ \hline
MiniXception          & 16                        & \begin{tabular}[c]{@{}l@{}}Stem widths are set to {[}w/2, w{]},\\ and block widths to {[}w, 2w, 4w, 8w{]}.\end{tabular}                                                \\ 
\bottomrule
\end{tabular}
\caption{Selected \textit{w} width parameter based on best performance for all models. Table and values obtained from \cite{Matias_Valdenegro_IV}.}
\label{table:rotnet_hyperparams_w}
\end{table}

\subsubsection{Self-Supervised Training on Watertank Dataset}

We trained our models on an NVIDIA Tesla V100 GPU (8GB of memory) using Keras and Tensorflow 2. Each  architecture was trained with the Adam optimizer \cite{Diederik_Kingma}, a learning rate $\beta = 0.001$ and a total of 200 epochs (except MobileNet with 220 epochs). All models were trained with sonar images of size $96 \times 96$. During training, we applied real-time random shifts $s_w$, $s_h$ to each image (horizontal and vertical), these shifts were sampled from a uniform distribution $s_w \sim \textit{U}(0, 0.1\textit{w})$ and $s_h \sim \textit{U}(0, 0.1\textit{h})$. Additionally, we applied random up-down and left-right flips with 50\% probability each one. We observed that normalizing the dataset by dividing pixel values by 255 causes unstable training, due to this, we restored to the mean subtraction normalization as performed in \cite{Matias_Valdenegro_IV}. The pixel mean value of the training set is $\mu_\text{pixel} = 84.5$ and normalization on each image is given by $x_\text{normalized} = x - \mu_\text{pixel}$.

\cref{table:rotnet_pretrain_accs} summarizes our training results on the Watertank dataset. The best performing model is \textit{ResNet20} followed by \textit{MiniXception}. We provide self-supervised classification accuracies (for rotation labels) and supervised classification accuracies (for actual class labels).
In this case, we have been able to apply standard supervised classification since we have the class labels from the Watertank objects. 

\begin{table}[]
\scriptsize
\centering
\begin{tabular}{lll}
\toprule
\textbf{Baseline Model}       & \textbf{Rotation Accuracy (SSL)} & \textbf{True label Accuracy (SL)} \\
\midrule
ResNet20     & 97.22\%  & 96.46\%  \\ \midrule
MobileNet    & 94.43\%  & 98.23\%  \\ \midrule
DenseNet121  & 95.38\%  & 96.46\%  \\ \midrule
SqueezeNet   & 95\%     & 97.47\%  \\ \midrule
Minixception & 96.14\%  & 96.71\%  \\ \midrule
Linear SVM   & 76.40\%  & 96.70\%  \\
\bottomrule
\end{tabular}
\caption{RotNet pre-training classification results on the test set of the Watertank dataset. Results are shown for all baseline CNN architectures.}
\label{table:rotnet_pretrain_accs}
\end{table}

\subsubsection{Transfer Learning on Turntable Dataset}

In order to evaluate the quality of the features learned by the models from the previous section, we performed transfer learning on the Turntable dataset. 
In this setup, we first subsampled the training set of the Turntable dataset to have samples-per-class (spc) $\in [10, 20, 30, 40, 50, 80, 110, 140, 170, 200]$ (to resemble a few-shot learning scheme). Thereafter, we generated embeddings of each subsampled training set using the pretrained models from above (we selected three hidden layers in each case close to the output, usually Flatten or last ReLU/Batch Normalization layers). Lastly, we used the embeddings to train a support vector machine classifier (SVM) with parameter  $C = 1.0$ (regularization parameter) for each spc (a total of 10 times to obtain a mean accuracy and standard deviation). Note that the SVM was not tuned though a cross-validation approach as the goal is to show the benefit SSL regardless of the performance of the model used for transfer learning.

We decided to use the linear SVM classifier as a first test of linear separability without having to perform further fine-tuning for transfer learning. We quantified sample complexity by recording classification accuracy on the test set of the Turntable dataset for each pretrained model, spc and hidden layer.

\begin{figure*}
  \centering
  \begin{subfigure}{0.3\linewidth}
  	\includegraphics[scale=0.26]{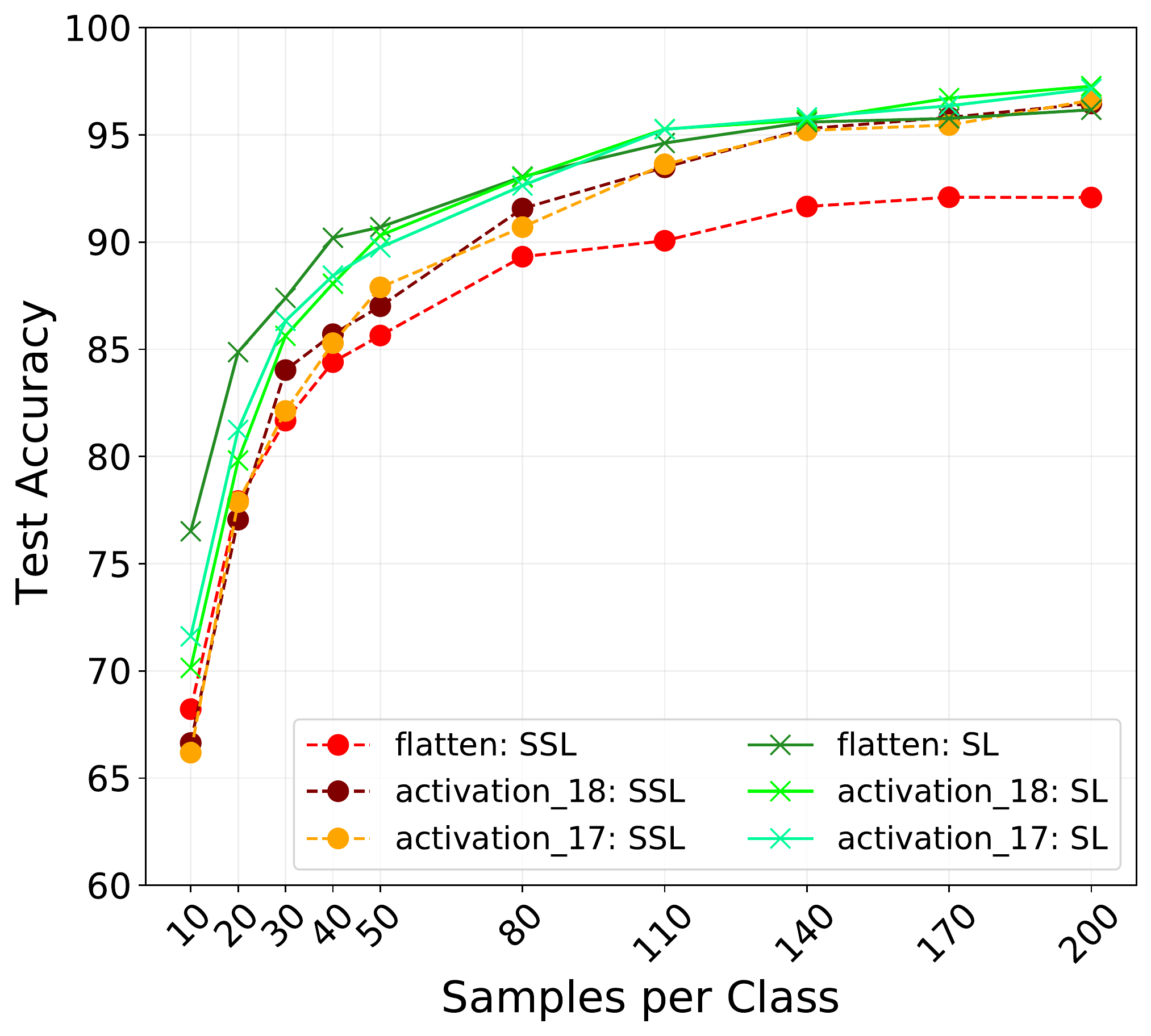}
    \caption{ResNet20}
    \label{fig:short-a}
  \end{subfigure}
  \hfill
  \begin{subfigure}{0.3\linewidth}
    \includegraphics[scale=0.26]{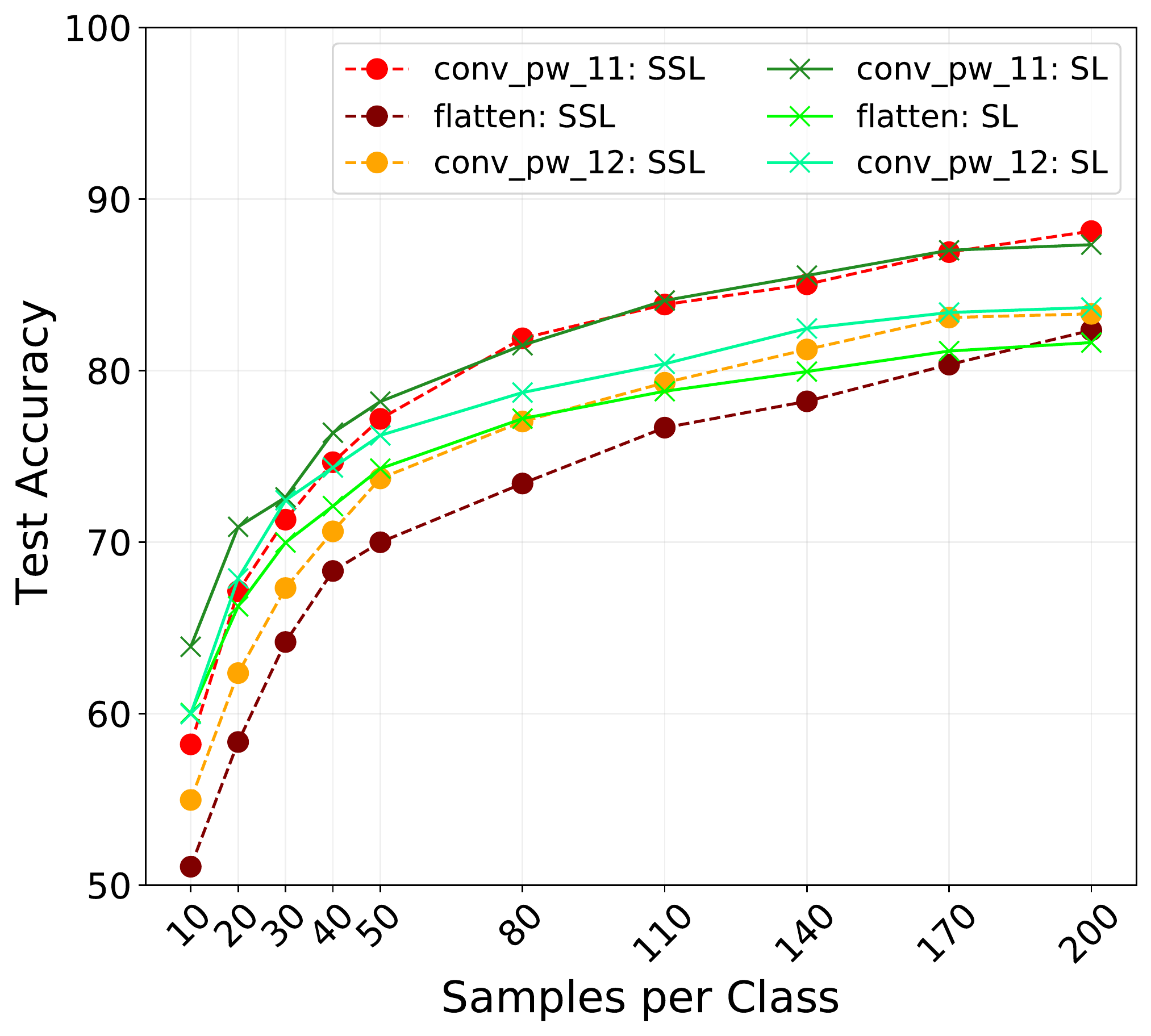}
    \caption{MobileNet}
    \label{fig:short-b}
  \end{subfigure}
  \hfill
  \begin{subfigure}{0.3\linewidth}
    \includegraphics[scale=0.26]{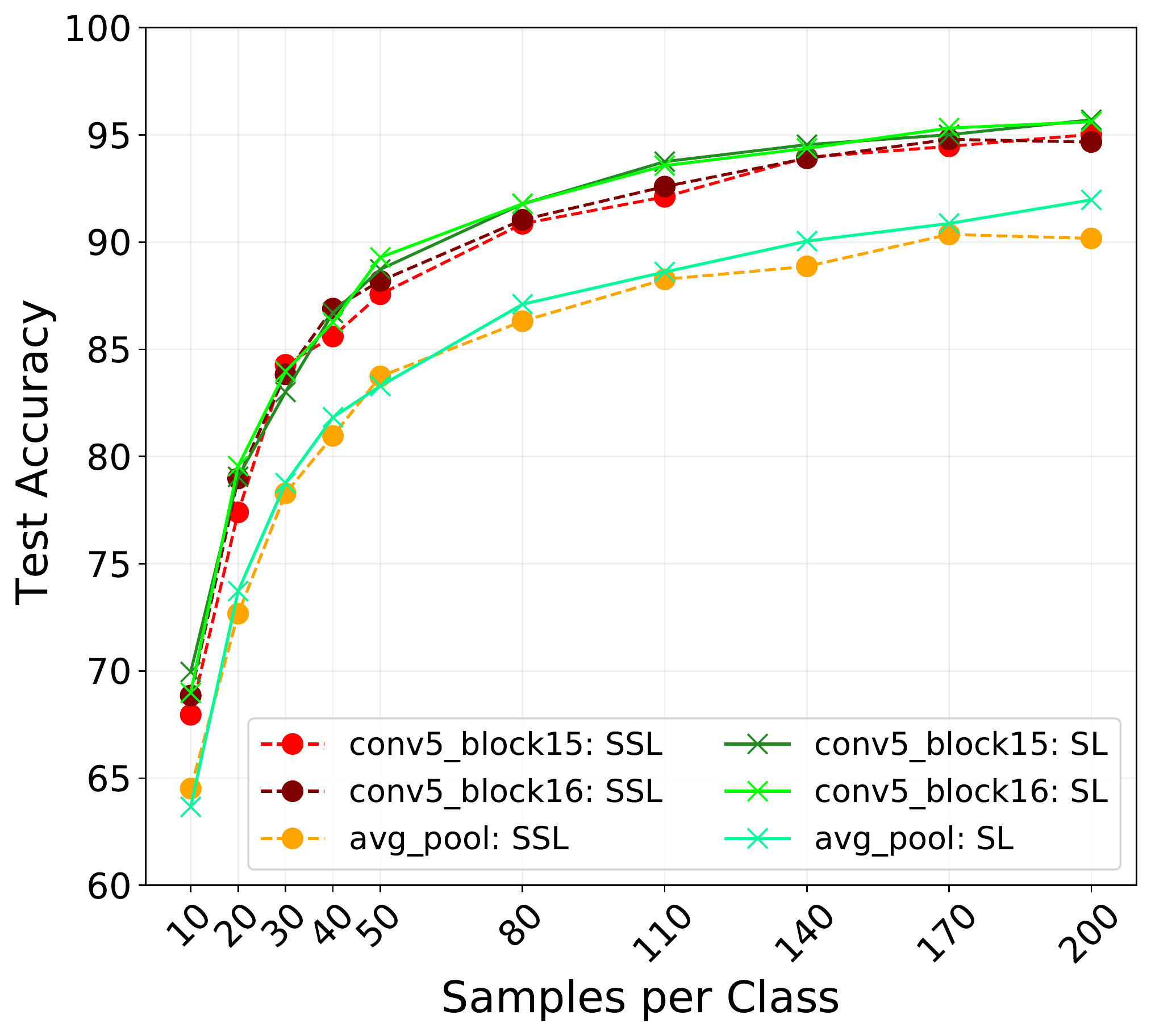}
    \caption{DenseNet121}
    \label{fig:short-b}
  \end{subfigure}

  \bigskip %
   \centering
  \begin{subfigure}{0.3\linewidth}
  	\includegraphics[scale=0.26]{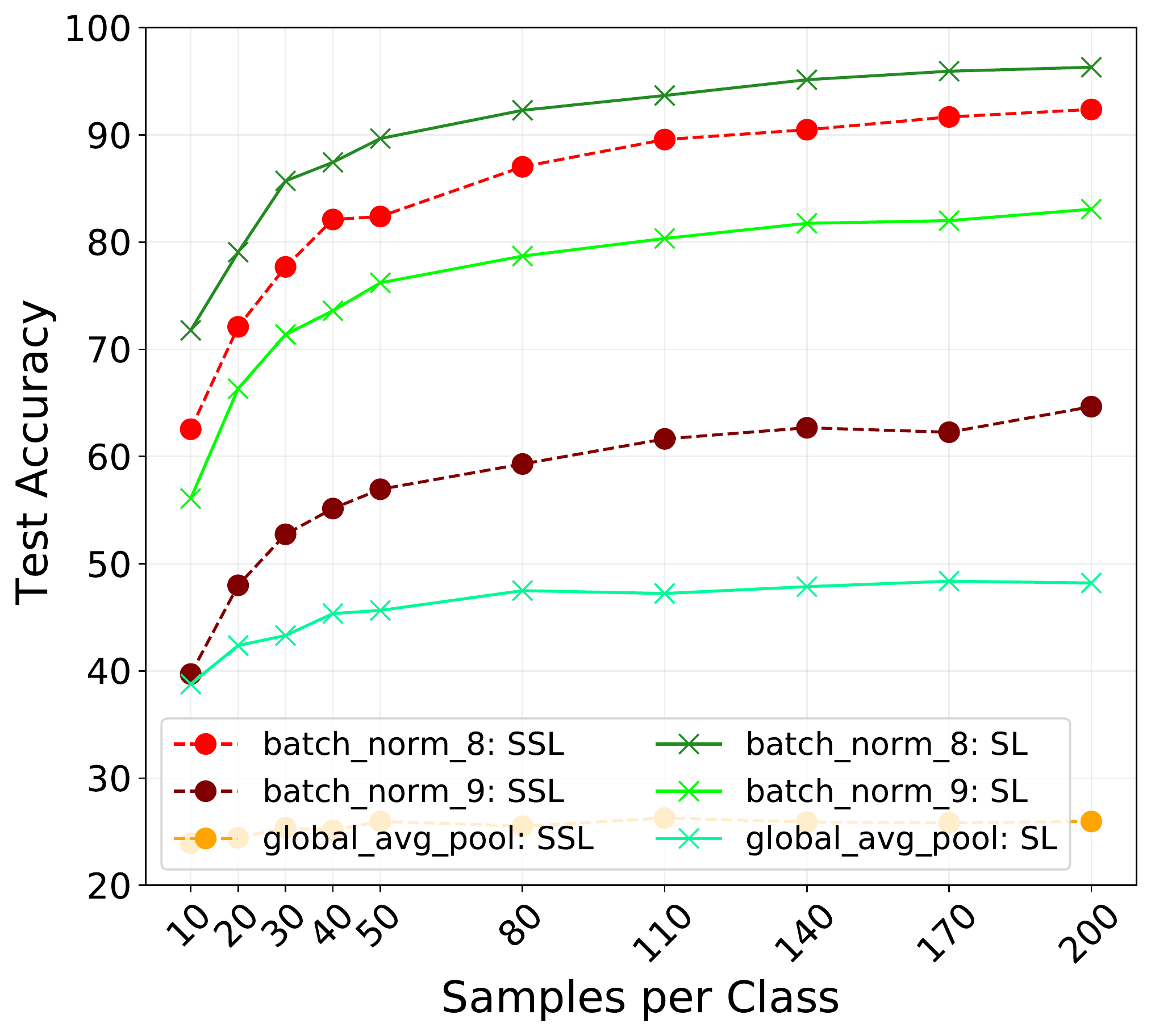}
    \caption{SqueezeNet}
    \label{fig:short-a}
  \end{subfigure}
  \hfill
  \begin{subfigure}{0.3\linewidth}
    \includegraphics[scale=0.26]{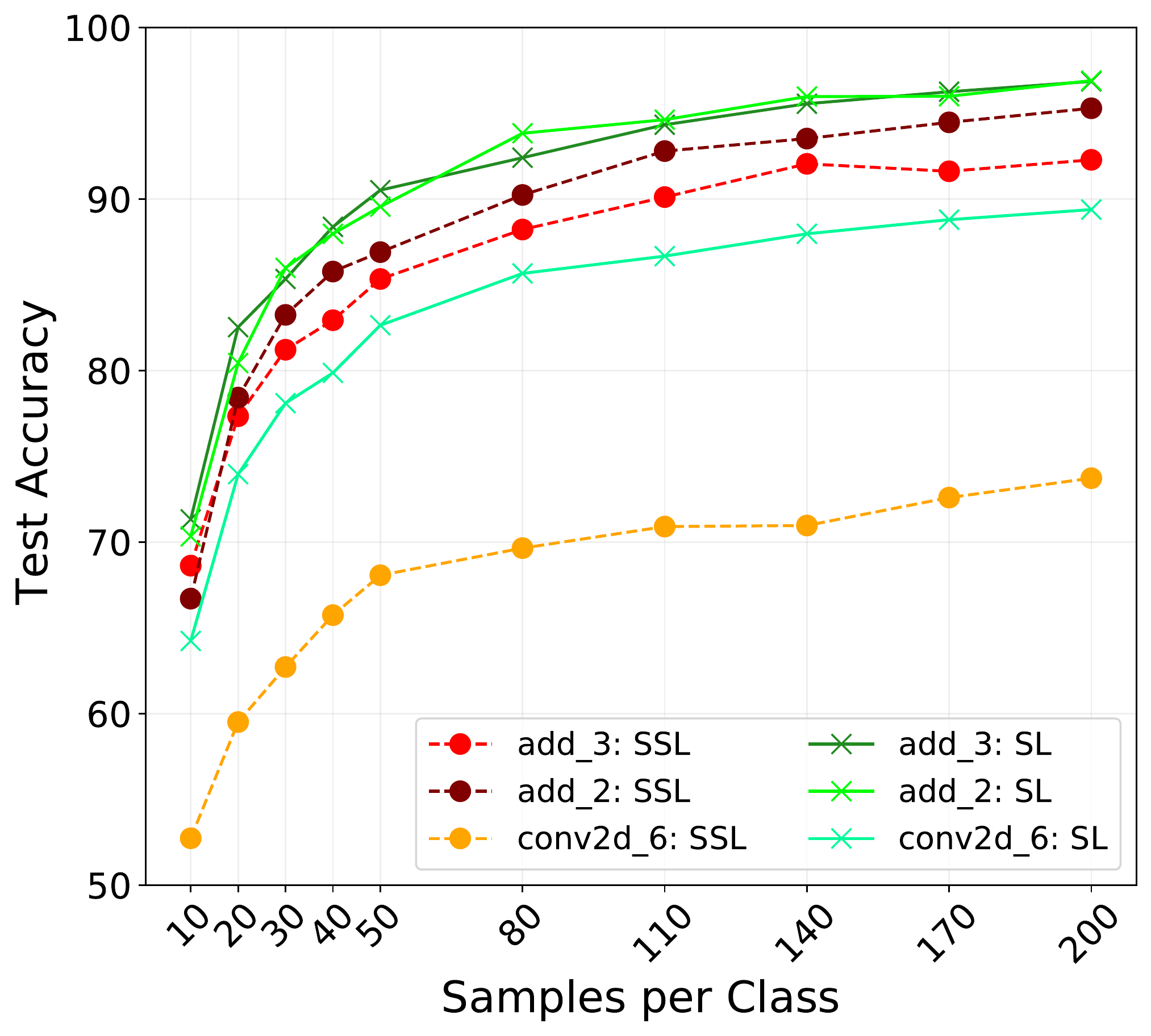}
    \caption{MiniXception}
    \label{fig:short-b}
  \end{subfigure}
  \hfill
  \begin{subfigure}{0.3\linewidth}
    \includegraphics[scale=0.26]{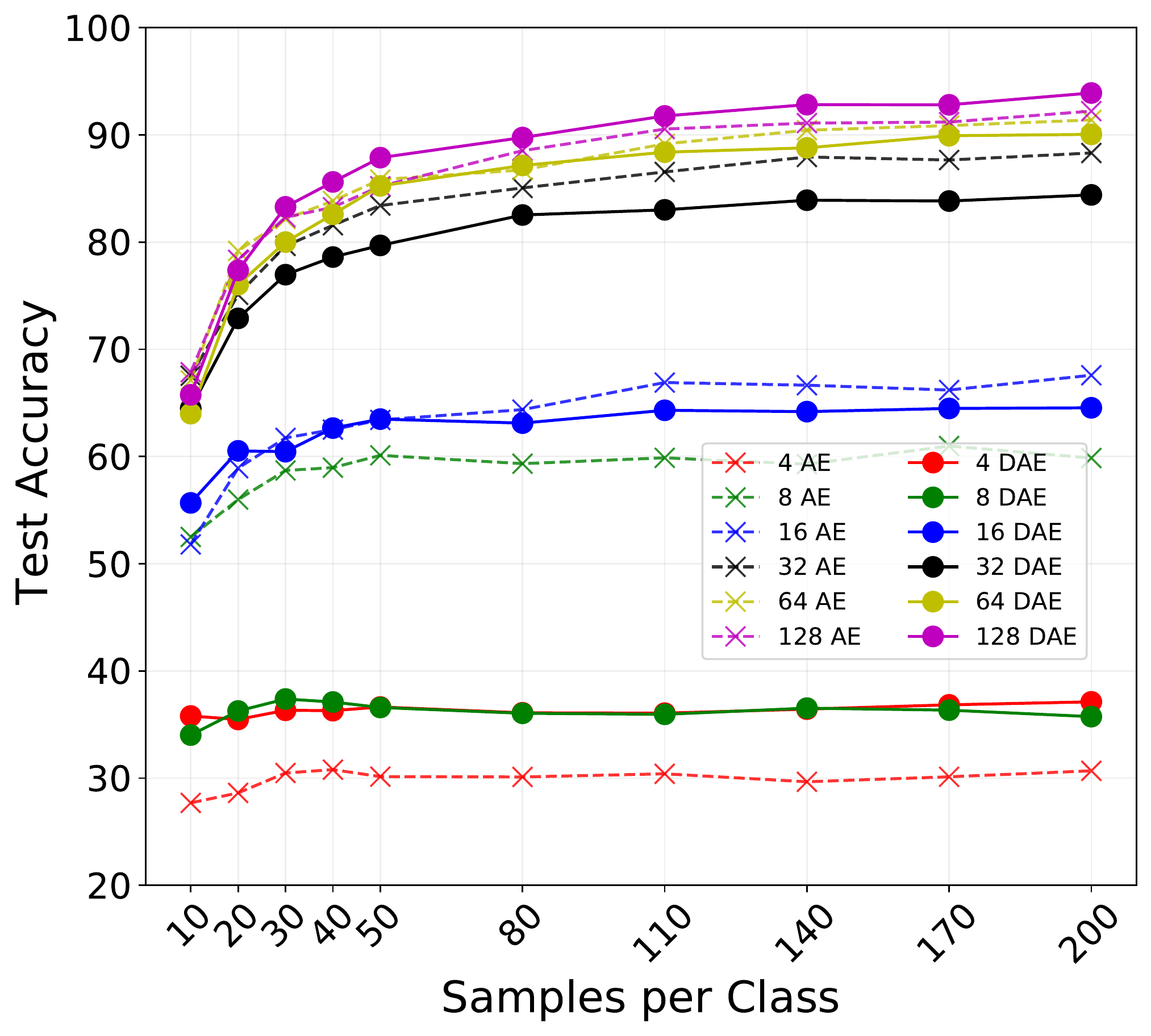}
    \caption{Denoising Autoencoder}
    \label{fig:short-b}
  \end{subfigure}
  \caption{Transfer learning results on the Turntable dataset for models pretrained on the Watertank dataset. Multiple selected hidden layers and number of samples per class are presented. Red lines: self-supervised pre-training, green lines: supervised pre-training.}
  \label{fig:rotnet_acc_plot}
\end{figure*}

Our transfer learning results are presented in \cref{fig:rotnet_acc_plot}(a-e). We show the accuracy curves for each model as the samples per each class in the Turntable dataset increase. The green lines correspond to supervised pre-training and the red lines correspond to self-supervised pre-training. We observe that the red (SSL) lines have a similar performance compared to the green lines (SL) in terms of classification accuracy. Furthermore, there are even cases where self-supervised pre-training is better than supervised pre-training (e.g. MobileNet with conv-pw-11-relu layer).

These results are an important first indicator that self-supervised pre-training can be a replacement for supervised pre-training for sonar image classification tasks, thus removing the necessity of labeling large sonar datasets manually. The best self-supervised performing model (based on accuracy) corresponds to \textbf{ResNet20, activation-17 layer} with an accuracy of \textbf{$96.62 \pm 0.562\%$}, which is less than 1\% difference compared to the best supervised performing model \textbf{ResNet20, activation-18 layer} with $97.27 ± 0.60\%$. Refer to \cref{tab:rotnet_tl_accs} in \cref{sec:detailed_rotnet_transfer_learning_results} for a detailed summary of all RotNet transfer learning evaluations. 

%

\subsection{Denoising Autoencoder - Learning Sonar Image Representations by Denoising Corrupted Inputs}

The Denoising Autoencoder (DAE) \cite{Pascal_I} applies Gaussian noise to input images and attempts to reconstruct the original uncorrupted inputs in an encoding-decoding manner. In the case of sonar images, this model is a good candidate to learn image representations while filtering noise out; this is of particular interest since most real-life sonar imaging scenarios consist of noisy images. Refer to \cref{sec:dae_model_selection} for a detailed description of the DAE architecture. In the following sections, we present pre-training and transfer learning experiments using the DAE algorithm. 


 \begin{figure}
		\centering
		\includegraphics[scale=0.2]{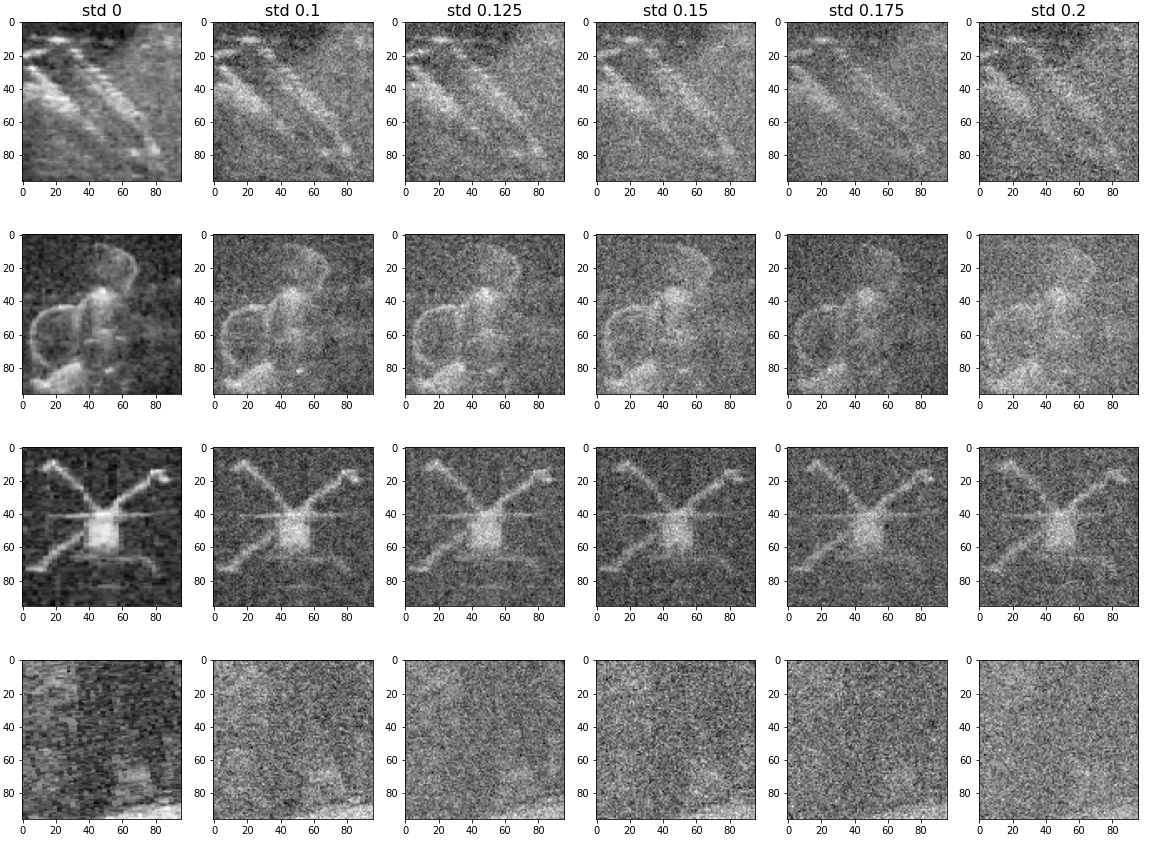}\hfill
		\caption{Different Gaussian noise standard deviation $\sigma$ applied to sonar images.}
		\label{fig:dae_noisy_samples}
		\end{figure}

%

\subsubsection{Self-Supervised Training on Watertank Dataset}

We trained the DAE with the standard Mean Squared Error (MSE) and Mean Average Error (MAE) loss metrics. Similar to RotNet, we implemented Adam optimizer, trained for 200 epochs, used batch sizes of 128, and a learning rate $\beta = 0.001$. Furthermore, we varied the code size (dimension of the intermediate layer) in the ranges $c \in [4, 8, 16, 32, 64, 128]$, with each value $c$ producing a different autoencoder model. This is a main parameter of the architecture as it can have an important impact on reconstruction performance. Gaussian noise was applied to each input image using Keras built-in \href{https://keras.io/api/layers/regularization_layers/gaussian_noise/}{Gaussian Noise Layer}. 
The main parameter of this noise layer is the standard deviation ($\sigma$) of the noise distribution as it controls the level of corruption applied to an image. During training, we have varied this value to $\sigma \in [0.100, 0.125, 0.150, 0.175, 0.200]$ to evaluate pre-training performance in the presence of different levels of noise (see \cref{fig:dae_noisy_samples}). 
In \cref{table:dae_pre-training_recs} we show the reconstruction results on the test set of the Watertank dataset for varying code sizes $c$ and $\sigma$. 
From these experiments, we observed that high code sizes ($c = 64, 128$) and lower $\sigma$ (due to less corruption) yield better reconstructions.


\begin{table}[]
\scriptsize
\centering
\begin{tabular}{llllll}
\toprule
 & & \multicolumn{4}{c}{Noise Standard Deviation $\mathbf{\sigma}$}\\
\textbf{Code Size} & \textbf{Metric}  & $\mathbf{0.125}$ & $\mathbf{0.150}$ & $\mathbf{0.175}$  & $\mathbf{0.200}$ \\
\midrule
\multirow{2}{*}{4}     & MSE  & 0.04 & 0.05 & 0.05 & 0.03      \\ 
                       & MAE  & 0.17 & 0.19 & 0.20 & 0.13           \\ \midrule
\multirow{2}{*}{8}    & MSE   & 0.05 & 0.04 & 0.03 & 0.05            \\  				  & MAE   & 0.19 & 0.18 & 0.13 & 0.19                  
\\ \midrule
\multirow{2}{*}{16}    & MSE   & 0.04 & 0.05 & 0.07 & 0.06            \\  				  & MAE   & 0.17 & 0.20 & 0.24 & 0.21                  
\\ \midrule
\multirow{2}{*}{32}    & MSE   & 0.03 & 0.05 & 0.04 & 0.04            \\  				  & MAE   & 0.15 & 0.20 & 0.19 & 0.18
\\ \midrule
\multirow{2}{*}{64}    & MSE   & 0.03 & 0.03 & 0.03 & 0.04
\\  				  & MAE   & 0.15 & 0.16 & 0.16 & 0.18                                    
\\ \midrule
\multirow{2}{*}{128}    & MSE   & \textbf{0.02} & 0.03 & 0.03 & 0.04            \\  	& MAE   & \textbf{0.12} & 0.16 & 0.14 & 0.17                  
\\
\bottomrule
\end{tabular}
\caption{DAE pre-training reconstruction results on the test set of the Watertank dataset. Results are shown for varying code sizes and standard deviations.}
\label{table:dae_pre-training_recs}
\end{table}

\subsubsection{Transfer Learning on Turntable Dataset}

Following a similar procedure as with RotNet, we have generated image embeddings with the pretrained DAE models and trained an SVM classifier by sub-sampling the training set of the Turntable dataset. In this case, we only used the encoder as a feature extractor for embedding generation, i.e. the encoder maps the input images of size $96 \times 96$ into a $1 \times c$ vector embedding for each code size $c$.

In \cref{fig:rotnet_acc_plot}(f), we show the transfer learning results for one DAE model ($\sigma = 0.1$) and an Auto Encoder (AE) (no denoising) and compare their performances across multiple code sizes $c$. We observed that there is a considerable gap in accuracies between DAE and AE for code size 4 (red lines). However, as we increased the code size, accuracies increase and the gap between both models decreases, indicating that higher dimensions can encode the original images more efficiently. For code size 128 for example, DAE performed better than AE, this indicates that adding noise as a form of random data augmentation can improve model performance. In this case, the best performing DAE (code size 128, $\sigma = 0.150$) obtained an accuracy of $94.49 \pm 0.62\%$, more than 2\% over the best performing AE with $92.21 \pm 0.59\%$. For a complete summary of our transfer learning experiments, refer to \cref{tab:dae_tl_accs} in \cref{sec:detailed_dae_transfer_learning_results}, which contains results for the best performing code sizes and all standard deviations.

\subsection{Jigsaw Puzzle - Learning Sonar Image Representations by Solving Jigsaw Puzzles}

The Jigsaw algorithm \cite{Noroozi_I} generates shuffled patches from input images as a form of self-supervision. First, an image is split into a $n \times n$ grid, yielding a total of $n^2$ image patches. Thereafter, the patches are shuffled based on predefined  permutation sets. Although there are $n!$ possible permutation sets, this can be narrowed down as desired (based on memory limitations). The goal of this algorithm is to classify which permutation set has been applied to a given input image. Refer to \cref{sec:jigsaw_model_selection} for a description of the Jigsaw architecture implemented. In the following sections, we describe technical aspects related to data generation, pre-training, and transfer learning evaluations of Jigsaw.

\subsubsection{Puzzle Data Generation}

\begin{figure}
		\centering
		\includegraphics[scale=0.65]{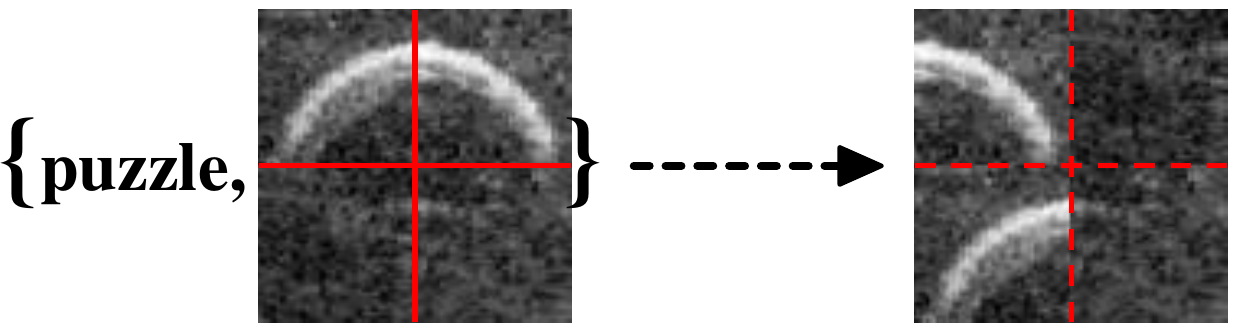}\hfill
		\caption{Jigsaw puzzle data generation. From left to right: grid generation and shuffling of patches ($2\times2$ example).}
		\label{fig:jigsaw_arranged_samples}
		\end{figure}

In order to generate jigsaw puzzles from the Watertank sonar dataset, first, we split them into a $3 \times 3$ grid. Thereafter, we shuffle the resulting image patches based on a randomly generated permutation set, since there are $9! = 362880$ possible ways to shuffle a $3 \times 3$ image grid, we narrow down the possible permutation sets to $\in [5, 10, 15, 20]$. \cref{fig:jigsaw_arranged_samples} shows a diagram with the described jigsaw data generation.


To frame the Jigsaw algorithm as a classification problem, we associate an integer label to each permutation set. In this sense, each image is shuffled according to a given permutation order with a deterministic integer label associated to it. 

\subsubsection{Self-Supervised Training on Watertank Dataset}

We have trained the Jigsaw model on the Watertank dataset, the original images are of size $96 \times 96$. After splitting them into $3\times3$ girds, the resulting patches are of size $32 \times 32$. We used a categorical cross-entropy loss for classification, Adam optimizer, learning rate $\beta = 0.001$,  batches of size 128 during 20 epochs (considerably less than RotNet and DAE). We trained the same architecture for four permutation sets $\{5, 10, 15, 20\}$ applied to the dataset separately. We summarize the pre-training results in \cref{table:jigsaw_pre-training_accs}. Notice that the original dimensions of the dataset increase by the permutation set size. In this case, all permutation sets yield similar test accuracies, with 5 permutations being the best performing one due to the fact that this is the minimum number of permutation sets the model has to classify.



\begin{table}[]
\scriptsize
\centering
\begin{tabular}{lll}
\toprule
\textbf{Permutations}       & \textbf{Test Set Size} & \textbf{Test Set Accuracy} \\
\midrule
\multirow{1}{*}{5}     & 1975      & \textbf{97.22}\%   
\\ \midrule
\multirow{1}{*}{10}    & 3950      & 96.76\% 
\\ \midrule
\multirow{1}{*}{15}    & 5925      & 96.03\%        
\\ \midrule
\multirow{1}{*}{20}   & 7900      & 94.87\%                   
\\
\bottomrule
\end{tabular}
\caption{Jigsaw pre-training classification results on the test set
of the Watertank dataset. Results are shown for different permutation sets.}
\label{table:jigsaw_pre-training_accs}
\end{table}

\subsubsection{Transfer Learning on Turntable Dataset}

For transfer learning evaluations on the Turntable dataset, we have taken the learned weights from the feature extractor pretrained with Jigsaw. For this integration, we had to change the input size of the feature extractor to take inputs of shape $96 \times 96$ corresponding to the actual image dimensions from the Turntable dataset, as the original Jigsaw model processes inputs with shapes $(9, 32, 32, 1)$ due to the 9 patches split. 
Similar to RotNet and DAE, we generate embeddings for the four pretrained Jigsaw models, one for each permutation set, and subsample the training set to train an SVM classifier with parameter $C = 1.0$.

\begin{figure*}
  \centering
  \begin{subfigure}{0.22\linewidth}
  	\includegraphics[scale=0.2]{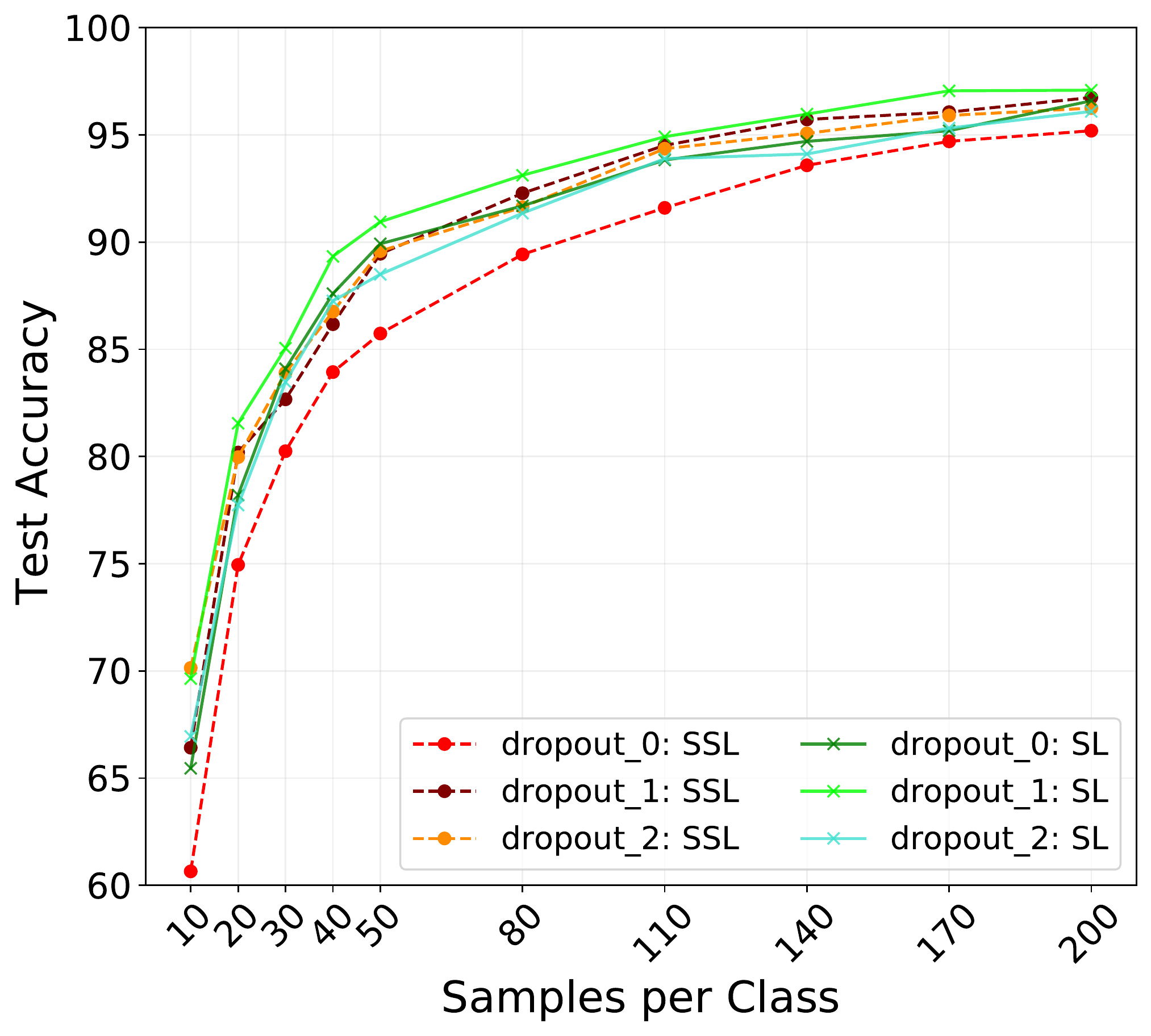}
    \caption{5 Permutations}
    \label{fig:short-a}
  \end{subfigure}
  \hfill
  \begin{subfigure}{0.22\linewidth}
    \includegraphics[scale=0.2]{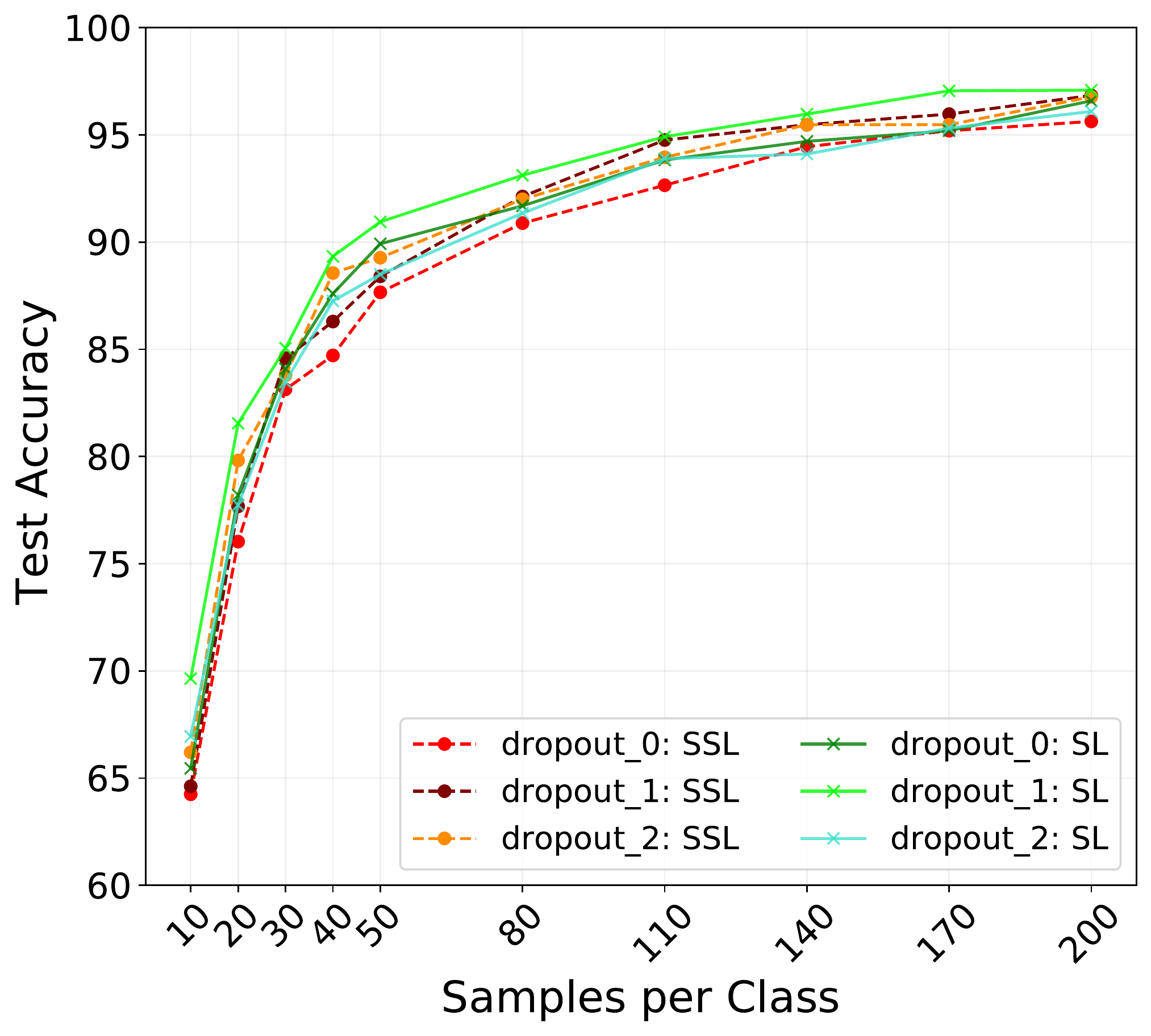}
    \caption{10 Permutations}
    \label{fig:short-b}
  \end{subfigure}
  \hfill
  \begin{subfigure}{0.22\linewidth}
  	\includegraphics[scale=0.2]{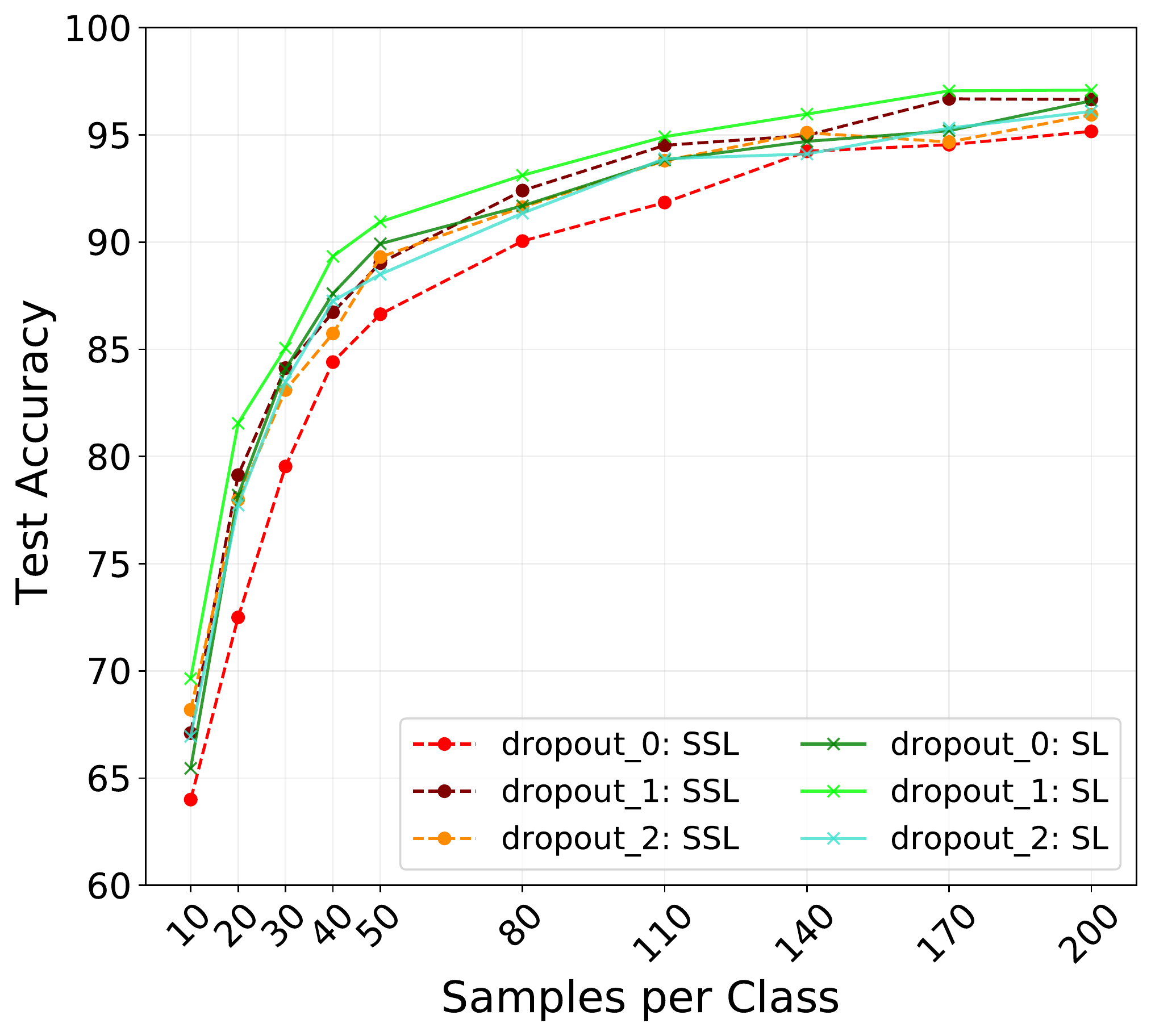}
    \caption{15 Permutations}
    \label{fig:short-a}
  \end{subfigure}
  \hfill
  \begin{subfigure}{0.22\linewidth}
    \includegraphics[scale=0.2]{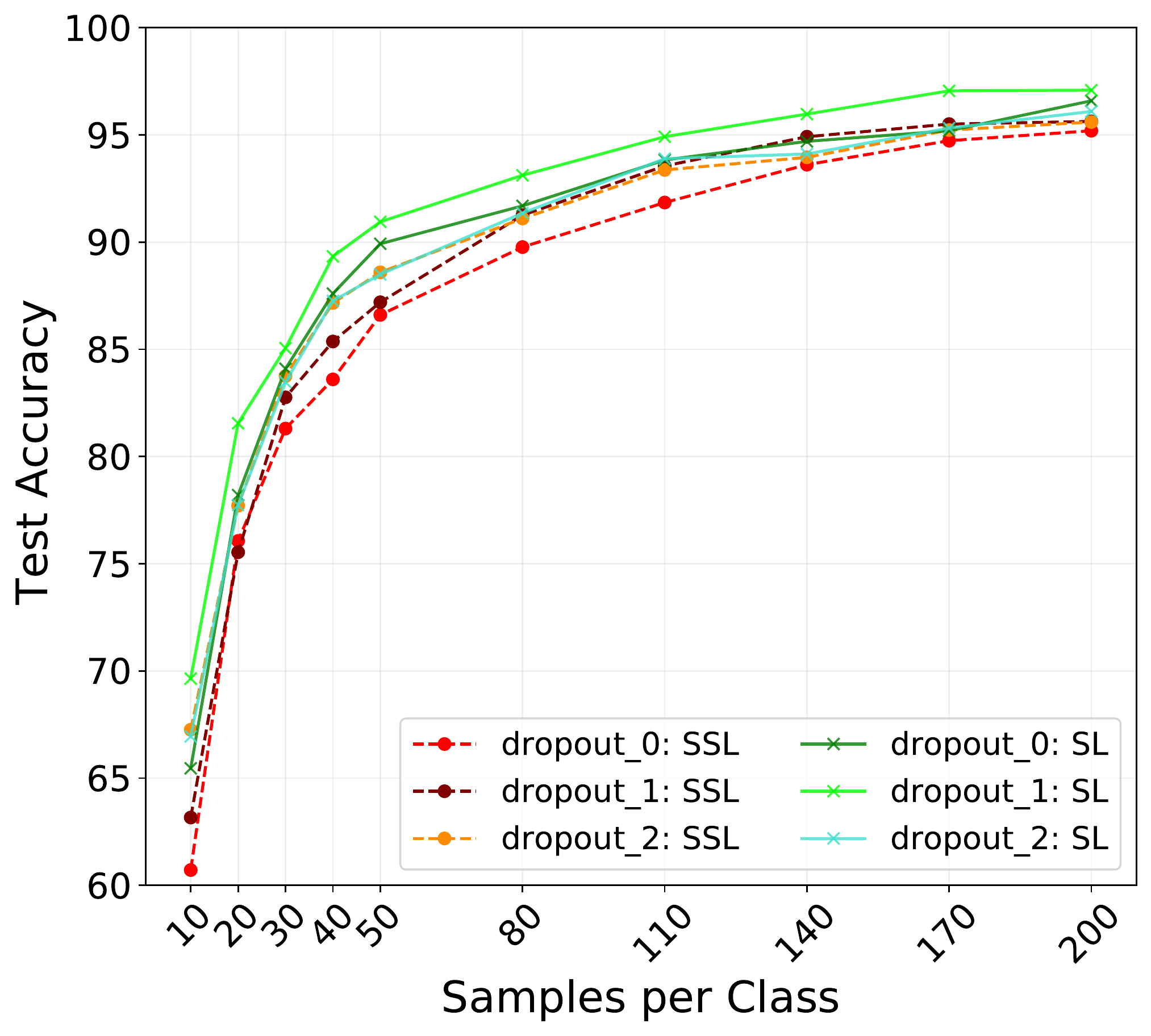}
    \caption{20 Permutations}
    \label{fig:short-b}
  \end{subfigure}
  \caption{Transfer learning evaluation on the turntable dataset with the Jigsaw model. Test accuracies are reported for varying number of permutations applied to each image in the dataset.}
  \label{fig:jigsaw_acc_plot}
\end{figure*}

In \cref{fig:jigsaw_acc_plot} we show the transfer learning for low-shot classification results on each permutation set. Similar to RotNet, we have taken three intermediate layers and evaluated their performance. Overall, the performance of the self-supervised pre-training curves is on par with the supervised pre-training curves. Similar to RotNet and DAE, lower samples per class (e.g. 10, 20, 30) yield higher variations, whereas higher samples per class have lower standard deviations and make training more stable. We have observed that the best performing Jigsaw model occurs with \textbf{10 permutations, dropout-1 layer} with an accuracy of $96.83 \pm 0.47 \%$, this is less than 1\% difference compared to the best supervised performing model with \textbf{dropout-1 layer} and accuracy of $97.08 \pm 0.22 \%$,. In \cref{tab:jigsaw_tl_accs} from \cref{sec:detailed_jigsaw_transfer_learning_results} we summarize the results for all permutation sets together with the supervised learning counterpart. 


\subsection{Self-Supervised Training on Wild Images}

\begin{figure}
  \centering
  \begin{subfigure}{0.45\linewidth}
  	\includegraphics[scale=0.45]{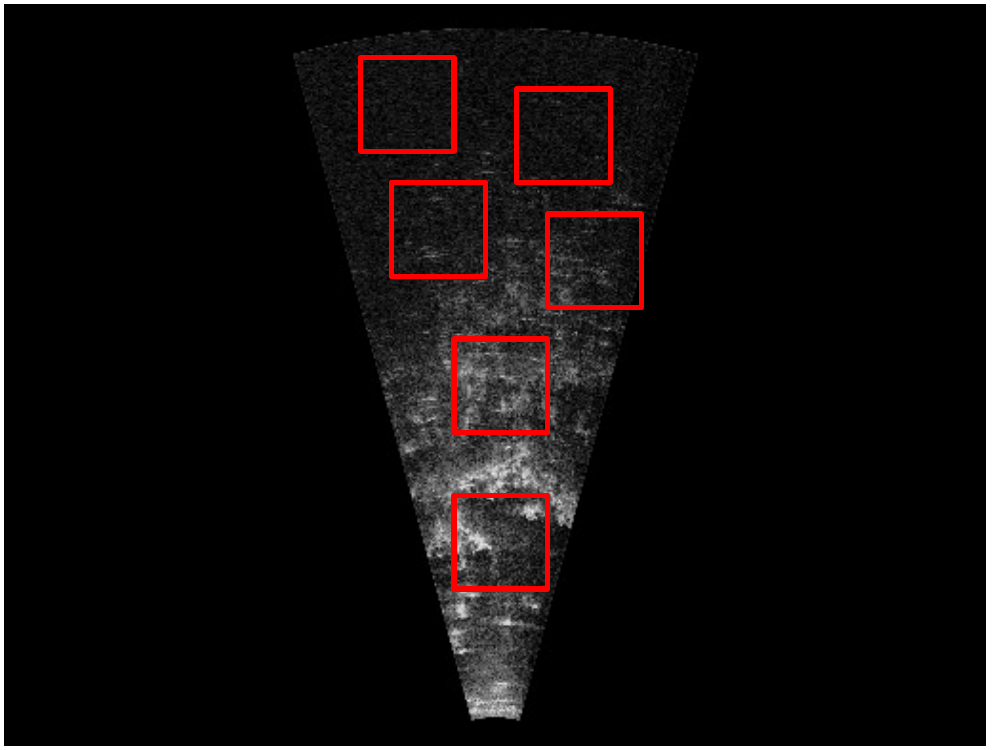}
    \caption{}
    \label{fig:short-a}
  \end{subfigure}
  \hfill
  \begin{subfigure}{0.4\linewidth}
    \includegraphics[scale=0.49]{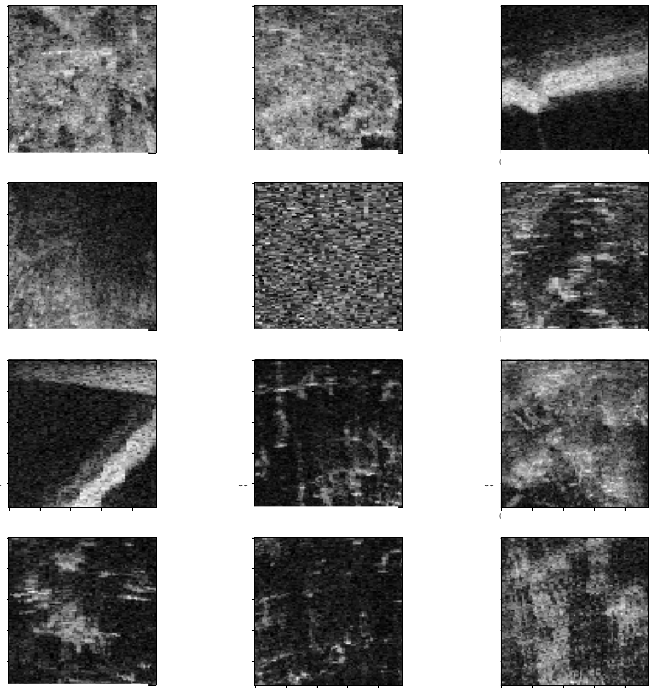}
    \caption{}
    \label{fig:short-b}
  \end{subfigure}
  \caption{Random patch generation from seabed sonar images: (a) patch region proposal, (b) resulting "wild" patches.}
  \label{fig:wild_data}
\end{figure}

In addition to the pre-training studies on the Watertank dataset carried out before, we have also pretrained our models on ``wild`'' uncurated sonar images. As discussed in \cite{Priya_Goyal_II}, computer vision models can be more robust when pretrained on uncurated images, one of the main reasons being that there are no well-defined classes and object shapes which can lead to biased results. Furthermore, characterizing the performance of our models on wild data is crucial in order to relax the constrain of having to manually annotate sonar datasets. For this, we collected sonar images from the seabed of a lake and generated random patches out of them. \cref{fig:wild_data} shows the proposed regions and the extracted patches. In total, we have generated 63,000 patches that contain different kinds of seabed shapes but no well-defined objects or shapes. We split the data into 80\% training, 20\% testing and trained our three proposed SSL methods with the same corresponding parameters described in the previous sections. Since this dataset has no class labels, we only trained with the synthetic labels from each self-supervised approach.

\begin{figure}
\centering

\includegraphics[scale=0.165]{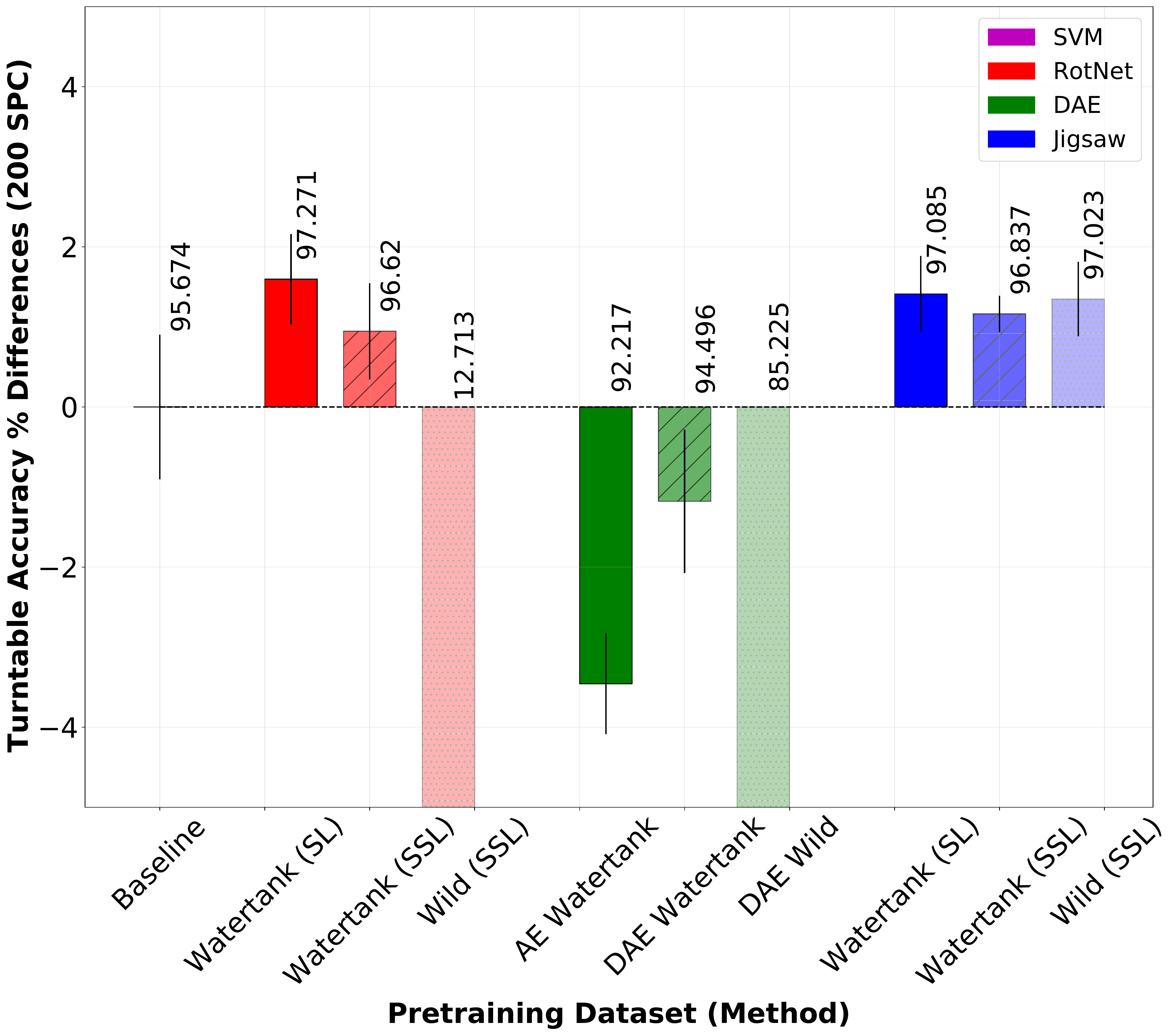}
\label{fig:200spc_bars_substract}
\caption{Comparison of baseline SVM, the best performing SSL models pretrained on the Watertank sonar dataset (with supervision and self-supervision) and the Wild sonar dataset. 200 spc case.}
\label{fig:200_spc_bars_full}
\end{figure}

In \cref{fig:200_spc_bars_full} and \cref{table:SSL_accs} we summarize the best performing models for Wild and Watertank data pre-training, evaluated on the Turntable dataset (200 spc case). 
We observed that RotNet Wild (SSL) has an overwhelming poor performance due to the fact that the model struggles at identifying rotations from wild objects that have no well-defined form and shape. Aside from this approach, most methods with Wild pre-training have a competitive performance against Watertank pre-training (for SL and SSL). Remarkably, Jigsaw (blue bars) with Wild pre-training performed on pair with supervised pre-training, thus showing its potential for broader real-life scenarios and paving the way to tackle further tasks with sonar images and SSL approaches. In \cref{fig:best_performing_bars_TL} from \cref{sec:detailed_best_transfer_learning_results} we show further model comparisons for the case of 10, 40 and 110 spc, the behavior in these cases is similar to 200 spc.

\section{Discussion and Analysis}

Across all three SSL methods, we found that, as opposed to the spc reported in \cite{Matias_Valdenegro_IV}, we have required up to 200 spc (all the dataset) in order to achieve accuracies higher than 90\%. This has to do with the fact that our pre-training is done on the watertank data and our transfer learning evaluation is done on the turntable data (the opposite direction), the reason behind this can be that the turntable dataset is statistically more diverse due to the fact that rotating objects can result in slightly different sonar images due to shadows and thus harder to classify.

We observe that improvement of self-supervised pre-training is more prominent for low-shot scenarios. For example, the Jigsaw Watertank SSL and Wild SSL has an increased accuracy from the baseline by around 7\% for 10 spc. As we increase these samples, the variation decreases since the baseline learns from more examples. Compared to RotNet and Jigsaw, DAE achieved considerably lower accuracies, we hypothesize that this is mainly due to the fact that this method does not increase the dataset dimensions (e.g. RotNet increases it by 4), but rather, the dataset corrupts only original images. Overall, the best performing model is Jigsaw, requiring only 20 epochs to reach competitive accuracies (against 200 epochs for RotNet and DAE). On top of this, the feature extractor used in Jigsaw has $\sim$6,000 parameters, whereas RotNet and DAE contain $\sim$1,000,000 and $\sim$200,000 parameters, respectively.

Overall we believe our results show that SSL is a viable alternative to pre-train neural networks for sonar image classification, but unlike many results in color images, SSL does not seem to improve over feature learning using supervised learning. It is encouraging for the marine robotics community that SSL can be used for pre-training, and we expect that our models and methods will be used in practice. We will release code and all self-supervised pre-trained models publicly once the paper is accepted.

\begin{table}[!t]
	\scriptsize
	\centering
	\begin{tabular}{lll}
		\toprule
		\textbf{Model}       & \textbf{Accuracy} & \textbf{Difference to Baseline} \\
		\midrule
		Baseline SVM           & $95.67\%$  & --  \\ \midrule
		Rotnet - Watertank SL  & $97.27\%$  & $+1.60\%$ \\
		Rotnet - Watertank SSL & $96.62\%$  & $+0.95\%$   \\
		Rotnet - Wild SSL      & $12.71\%$  & $-82.96\%$  \\ \midrule
		AE - Watertank SL      & $92.22\%$  & $-3.45\%$  \\
		DAE - Watertank SSL    & $94.50\%$  & $-1.17\%$   \\
		DAE - Wild SSL         & $85.23\%$  & $-10.44\%$  \\ \midrule
		Jigsaw - Watertank SL  & $97.09\%$  & $+1.42\%$  \\
		Jigsaw - Watertank SSL & $96.84\%$  & $+1.17\%$   \\
		Jigsaw - Wild SSL      & $97.02\%$  & $+1.35\%$  \\ 
		\bottomrule
	\end{tabular}
	\caption{Comparison of the baseline SVM, the pre-trained models on the Watertank sonar dataset and the Wild sonar dataset.}
	\label{table:SSL_accs}
\end{table}
\section{Conclusions and Future Work}
\label{sec:conclusions}


In this work we evaluate three self-supervised learning algorithms for learning features from sonar images, using two datasets (marine debris turntable and on the wild sonar image patches), and evaluated them in classifying the marine debris watertank datset. These results prove that SSL methods that are commonly used for RBG images also work for sonar images and perform similarly.

Our results show that the performance of the evaluated self-supervised algorithms is on par with supervised algorithms for sonar image classification. We have verified experimentally that our algorithms can generate high-quality image embeddings for classification after a pre-training and transfer learning procedure. Across all three algorithms, the differences between SSL and SL in the obtained accuracies are small (only 1-2\%), this positions self-supervised learning as a strong candidate in the absence of labeled data. We notice that Jigsaw is the best performing SSL algorithm with a best accuracy of 97.02\% (surpassing RotNet with 96.46\% and DAE with 94.49\%), we argue that this is because this method benefits from image augmentation (synthetic labels) by design, remarkably, this result was achieved by pre-training on the wild uncurated sonar dataset, proving the potential of SSL to avoid labeling large sonar datasets (saving time and costs) and learning image representations from a diverse set of object shapes and forms.	

We expect that our results increase the use of self-supervised learning for feature learning in sonar images, improving the overall autonomy and perception capabilities of underwater vehicles. 

For future work, we plan to further investigate Transformer-based architectures for pre-training. We also plan to deploy some of our best performing pretrained models in real-life underwater environments for object classification. Lastly, we want to investigate further SSL techniques for the problem of image translation using multiple learning modalities (sonar and camera data).

\section*{Acknowledgements}
This work was done within the project DeeperSense that received funding from European Commision H2020-ICT-2020-2. Project Number: 101016958.

 \clearpage
{\small
    \bibliographystyle{ieee_fullname}
    \bibliography{egbib}

\begin{thebibliography}{10}\itemsep=-1pt

\bibitem{arriaga2017real}
Octavio Arriaga, Matias Valdenegro-Toro, and Paul Pl{\"o}ger.
\newblock Real-time convolutional neural networks for emotion and gender
  classification.
\newblock {\em arXiv preprint arXiv:1710.07557}, 2017.

\bibitem{Mathilde_Caron}
Mathilde Caron, Ishan Misra, Julien Mairal, Priya Goyal, Piotr Bojanowski, and
  Armand Joulin.
\newblock Unsupervised learning of visual features by contrasting cluster
  assignments.
\newblock In H. Larochelle, M. Ranzato, R. Hadsell, M.~F. Balcan, and H. Lin,
  editors, {\em Advances in Neural Information Processing Systems}, volume~33,
  pages 9912--9924. Curran Associates, Inc., 2020.

\bibitem{Ting_Chen}
Ting Chen, Simon Kornblith, Mohammad Norouzi, and Geoffrey Hinton.
\newblock A simple framework for contrastive learning of visual
  representations.
\newblock In Hal~Daumé III and Aarti Singh, editors, {\em Proceedings of the
  37th International Conference on Machine Learning}, volume 119 of {\em
  Proceedings of Machine Learning Research}, pages 1597--1607. PMLR, 13--18 Jul
  2020.

\bibitem{Francois_Chollet}
Francois Chollet.
\newblock Xception: Deep learning with depthwise separable convolutions.
\newblock {\em 2017 IEEE Conference on Computer Vision and Pattern Recognition
  (CVPR)}, pages 1800--1807, 2017.

\bibitem{Zeyu_Feng}
Zeyu Feng, Chang Xu, and Dacheng Tao.
\newblock Self-supervised representation learning by rotation feature
  decoupling.
\newblock In {\em Proceedings of the IEEE/CVF Conference on Computer Vision and
  Pattern Recognition (CVPR)}, June 2019.

\bibitem{Gidaris_I}
Spyros Gidaris, Praveer Singh, and Nikos Komodakis.
\newblock Unsupervised representation learning by predicting image rotations,
  2018.

\bibitem{Priya_Goyal_II}
Priya Goyal, Quentin Duval, Isaac Seessel, Mathilde Caron, Mannat Singh, Ishan
  Misra, Levent Sagun, Armand Joulin, and Piotr Bojanowski.
\newblock Vision models are more robust and fair when pretrained on uncurated
  images without supervision.
\newblock {\em arXiv preprint arXiv:2202.08360}, 2022.

\bibitem{Priya_goyal}
Priya Goyal, Dhruv Mahajan, Abhinav Gupta, and Ishan Misra.
\newblock Scaling and benchmarking self-supervised visual representation
  learning.
\newblock In {\em Proceedings of the IEEE/CVF International Conference on
  Computer Vision (ICCV)}, October 2019.

\bibitem{Jean_Grill}
Jean-Bastien Grill, Florian Strub, Florent Altch\'{e}, Corentin Tallec, Pierre
  Richemond, Elena Buchatskaya, Carl Doersch, Bernardo Avila~Pires, Zhaohan
  Guo, Mohammad Gheshlaghi~Azar, Bilal Piot, koray kavukcuoglu, Remi Munos, and
  Michal Valko.
\newblock Bootstrap your own latent - a new approach to self-supervised
  learning.
\newblock In H. Larochelle, M. Ranzato, R. Hadsell, M.~F. Balcan, and H. Lin,
  editors, {\em Advances in Neural Information Processing Systems}, volume~33,
  pages 21271--21284. Curran Associates, Inc., 2020.

\bibitem{Kaiming_He}
Kaiming He, Xiangyu Zhang, Shaoqing Ren, and Jian Sun.
\newblock Deep residual learning for image recognition.
\newblock In {\em Proceedings of the IEEE conference on computer vision and
  pattern recognition}, pages 770--778, 2016.

\bibitem{Olivier_Henaff}
Olivier Henaff.
\newblock Data-efficient image recognition with contrastive predictive coding.
\newblock In Hal~Daumé III and Aarti Singh, editors, {\em Proceedings of the
  37th International Conference on Machine Learning}, volume 119 of {\em
  Proceedings of Machine Learning Research}, pages 4182--4192. PMLR, 13--18 Jul
  2020.

\bibitem{Dan_Hendrycks}
Dan Hendrycks, Mantas Mazeika, Saurav Kadavath, and Dawn Song.
\newblock Using self-supervised learning can improve model robustness and
  uncertainty.
\newblock In {\em Advances in Neural Information Processing Systems},
  volume~32. Curran Associates, Inc., 2019.

\bibitem{Andrew_Howard}
Andrew~G Howard, Menglong Zhu, Bo Chen, Dmitry Kalenichenko, Weijun Wang,
  Tobias Weyand, Marco Andreetto, and Hartwig Adam.
\newblock Mobilenets: Efficient convolutional neural networks for mobile vision
  applications.
\newblock {\em arXiv preprint arXiv:1704.04861}, 2017.

\bibitem{Gao_Huang}
Gao Huang, Zhuang Liu, Laurens van~der Maaten, and Kilian~Q Weinberger.
\newblock Densely connected convolutional networks.
\newblock In {\em Proceedings of the IEEE Conference on Computer Vision and
  Pattern Recognition}, 2017.

\bibitem{Forrest_Iandola}
Forrest~N Iandola, Song Han, Matthew~W Moskewicz, Khalid Ashraf, William~J
  Dally, and Kurt Keutzer.
\newblock Squeezenet: Alexnet-level accuracy with 50x fewer parameters and< 0.5
  mb model size.
\newblock {\em arXiv preprint arXiv:1602.07360}, 2016.

\bibitem{Satoshi_I}
Satoshi Iizuka, Edgar Simo-Serra, and Hiroshi Ishikawa.
\newblock Globally and locally consistent image completion.
\newblock {\em ACM Trans. Graph.}, 36(4), July 2017.

\bibitem{jaiswal2021survey}
Ashish Jaiswal, Ashwin~Ramesh Babu, Mohammad~Zaki Zadeh, Debapriya Banerjee,
  and Fillia Makedon.
\newblock A survey on contrastive self-supervised learning.
\newblock {\em Technologies}, 9(1):2, 2021.

\bibitem{jang2019cnn}
Hyesu Jang, Giseop Kim, Yeongjun Lee, and Ayoung Kim.
\newblock Cnn-based approach for opti-acoustic reciprocal feature matching.
\newblock In {\em ICRA Workshop on Underwater Robotics Perception}, 2019.

\bibitem{Jing_2020}
Longlong Jing and Yingli Tian.
\newblock Self-supervised visual feature learning with deep neural networks: A
  survey.
\newblock {\em IEEE transactions on pattern analysis and machine intelligence},
  43(11):4037--4058, 2020.

\bibitem{kim2017denoising}
Juhwan Kim, Seokyong Song, and Son-Cheol Yu.
\newblock Denoising auto-encoder based image enhancement for high resolution
  sonar image.
\newblock In {\em 2017 IEEE underwater technology (UT)}, pages 1--5. IEEE,
  2017.

\bibitem{Diederik_Kingma}
Diederik Kingma and Jimmy Ba.
\newblock Adam: A method for stochastic optimization.
\newblock {\em International Conference on Learning Representations}, 12 2014.

\bibitem{Larsson_I}
Gustav Larsson, Michael Maire, and Gregory Shakhnarovich.
\newblock Colorization as a proxy task for visual understanding.
\newblock In {\em Proceedings of the IEEE conference on computer vision and
  pattern recognition}, pages 6874--6883, 2017.

\bibitem{Ledig_I}
Christian Ledig, Lucas Theis, Ferenc Husz{\'a}r, Jose Caballero, Andrew
  Cunningham, Alejandro Acosta, Andrew Aitken, Alykhan Tejani, Johannes Totz,
  Zehan Wang, et~al.
\newblock Photo-realistic single image super-resolution using a generative
  adversarial network.
\newblock In {\em Proceedings of the IEEE conference on computer vision and
  pattern recognition}, pages 4681--4690, 2017.

\bibitem{Lee_2017}
Hsin-Ying Lee, Jia-Bin Huang, Maneesh Singh, and Ming-Hsuan Yang.
\newblock Unsupervised representation learning by sorting sequences.
\newblock In {\em Proceedings of the IEEE international conference on computer
  vision}, pages 667--676, 2017.

\bibitem{Meng_Li}
Meng Li, William Hsu, Xiaodong Xie, Jason Cong, and Wen Gao.
\newblock Sacnn: Self-attention convolutional neural network for low-dose ct
  denoising with self-supervised perceptual loss network.
\newblock {\em IEEE Transactions on Medical Imaging}, 39(7):2289--2301, 2020.

\bibitem{Ru_Li}
Ru Li, Shuaicheng Liu, Guangfu Wang, Guanghui Liu, and Bing Zeng.
\newblock Jigsawgan: Auxiliary learning for solving jigsaw puzzles with
  generative adversarial networks.
\newblock {\em IEEE Transactions on Image Processing}, 31:513--524, 2021.

\bibitem{liu2021self}
Xiao Liu, Fanjin Zhang, Zhenyu Hou, Li Mian, Zhaoyu Wang, Jing Zhang, and Jie
  Tang.
\newblock Self-supervised learning: Generative or contrastive.
\newblock {\em IEEE Transactions on Knowledge and Data Engineering}, 2021.

\bibitem{Ishan_Misra}
Ishan Misra and Laurens van~der Maaten.
\newblock Self-supervised learning of pretext-invariant representations.
\newblock In {\em Proceedings of the IEEE/CVF Conference on Computer Vision and
  Pattern Recognition (CVPR)}, June 2020.

\bibitem{Misra_2016}
Ishan Misra, C~Lawrence Zitnick, and Martial Hebert.
\newblock Shuffle and learn: unsupervised learning using temporal order
  verification.
\newblock In {\em European Conference on Computer Vision}, pages 527--544.
  Springer, 2016.

\bibitem{Noroozi_I}
Mehdi Noroozi and Paolo Favaro.
\newblock Unsupervised learning of visual representations by solving jigsaw
  puzzles.
\newblock In {\em European conference on computer vision}, pages 69--84.
  Springer, 2016.

\bibitem{Pathak_I}
Deepak Pathak, Philipp Krahenbuhl, Jeff Donahue, Trevor Darrell, and Alexei~A
  Efros.
\newblock Context encoders: Feature learning by inpainting.
\newblock In {\em Proceedings of the IEEE conference on computer vision and
  pattern recognition}, pages 2536--2544, 2016.

\bibitem{saucan2015model}
Augustin-Alexandru Saucan, Christophe Sintes, Thierry Chonavel, and Jean-Marc
  Le~Caillec.
\newblock Model-based adaptive 3d sonar reconstruction in reverberating
  environments.
\newblock {\em IEEE transactions on image processing}, 24(10):2928--2940, 2015.

\bibitem{Srivastava_2015}
Nitish Srivastava, Elman Mansimov, and Ruslan Salakhudinov.
\newblock Unsupervised learning of video representations using lstms.
\newblock In {\em International conference on machine learning}, pages
  843--852. PMLR, 2015.

\bibitem{Vladimiros_Sterzentsenko}
Vladimiros Sterzentsenko, Leonidas Saroglou, Anargyros Chatzitofis, Spyridon
  Thermos, Nikolaos Zioulis, Alexandros Doumanoglou, Dimitrios Zarpalas, and
  Petros Daras.
\newblock Self-supervised deep depth denoising.
\newblock In {\em Proceedings of the IEEE/CVF International Conference on
  Computer Vision (ICCV)}, October 2019.

\bibitem{sung2018crosstalk}
Minsung Sung, Hyeonwoo Cho, Hangil Joe, Byeongjin Kim, and Son-Choel Yu.
\newblock Crosstalk noise detection and removal in multi-beam sonar images
  using convolutional neural network.
\newblock In {\em OCEANS 2018 MTS/IEEE Charleston}, pages 1--6, 2018.

\bibitem{Taleb_2021}
Aiham Taleb, Christoph Lippert, Tassilo Klein, and Moin Nabi.
\newblock Multimodal self-supervised learning for medical image analysis.
\newblock In {\em International Conference on Information Processing in Medical
  Imaging}, pages 661--673. Springer, 2021.

\bibitem{terayama2019integration}
Kei Terayama, Kento Shin, Katsunori Mizuno, and Koji Tsuda.
\newblock Integration of sonar and optical camera images using deep neural
  network for fish monitoring.
\newblock {\em Aquacultural Engineering}, 86:102000, 2019.

\bibitem{valdenegro2016object}
Matias Valdenegro-Toro.
\newblock Object recognition in forward-looking sonar images with convolutional
  neural networks.
\newblock In {\em OCEANS 2016 MTS/IEEE Monterey}, pages 1--6, 2016.

\bibitem{Matias_Valdenegro_I}
Matias Valdenegro-Toro.
\newblock Best practices in convolutional networks for forward-looking sonar
  image recognition.
\newblock In {\em OCEANS 2017-Aberdeen}, pages 1--9. IEEE, 2017.

\bibitem{Matias_Valdenegro_IV}
Matias Valdenegro-Toro, Alan Preciado-Grijalva, and Bilal Wehbe.
\newblock Pre-trained models for sonar images.
\newblock In {\em OCEANS 2021: San Diego – Porto}, pages 1--8, 2021.

\bibitem{Pascal_I}
Pascal Vincent, Hugo Larochelle, Yoshua Bengio, and Pierre-Antoine Manzagol.
\newblock Extracting and composing robust features with denoising autoencoders.
\newblock In {\em Proceedings of the 25th International Conference on Machine
  Learning}, ICML '08, page 1096–1103, New York, NY, USA, 2008. Association
  for Computing Machinery.

\bibitem{wang2019underwater}
Xingmei Wang, Jia Jiao, Jingwei Yin, Wensheng Zhao, Xiao Han, and Boxuan Sun.
\newblock Underwater sonar image classification using adaptive weights
  convolutional neural network.
\newblock {\em Applied Acoustics}, 146:145--154, 2019.

\bibitem{Yaochen_Xie}
Yaochen Xie, Zhengyang Wang, and Shuiwang Ji.
\newblock Noise2same: Optimizing a self-supervised bound for image denoising.
\newblock In H. Larochelle, M. Ranzato, R. Hadsell, M.~F. Balcan, and H. Lin,
  editors, {\em Advances in Neural Information Processing Systems}, volume~33,
  pages 20320--20330. Curran Associates, Inc., 2020.

\bibitem{Zhang_I}
Richard Zhang, Phillip Isola, and Alexei~A Efros.
\newblock Colorful image colorization.
\newblock In {\em European conference on computer vision}, pages 649--666.
  Springer, 2016.

\bibitem{Zhang_III}
Richard Zhang, Jun-Yan Zhu, Phillip Isola, Xinyang Geng, Angela~S. Lin, Tianhe
  Yu, and Alexei~A. Efros.
\newblock Real-time user-guided image colorization with learned deep priors.
\newblock {\em ACM Trans. Graph.}, 36(4), jul 2017.

\end{thebibliography}
}


\clearpage
\appendix
\FloatBarrier
\onecolumn

\section{Data and Models Release}

All training data used in this paper is available at  \url{https://github.com/mvaldenegro/marine-debris-fls-datasets/} and models and source code is available at \url{https://github.com/agrija9/ssl-sonar-images}.

\section{RotNet Model Selection}
\label{sec:rotnet_model_selection}

In this section we describe the neural network architectures that we used for evaluating RotNet.\\

Similar to the work in \cite{Matias_Valdenegro_IV}, we have implemented the several CNN architectures to evaluate the performance of RotNet in sonar images. All the described network architectures are lightweight versions that try to address memory and performance challenges in underwater robotics:


\textbf{ResNet \cite{Kaiming_He}:} Uses residual connections to improve gradient propagation through the network, this feature allows to design much deeper networks. For our experiments we used a compact variant (ResNet20) which has 20 residual layers.

\textbf{MobileNet \cite{Andrew_Howard}:} It is designed to be lightweight in order to do fast inference in mobile devices. It features depthwise separable convolutions to reduce computations (this induces a trade-off between accuracy performance and computation performance). 

\textbf{DenseNet \cite{Gao_Huang}:} It reuses features heavily. It is composed by a set of dense convolutional blocks, where each of them is a series of convolutional layers, and each layer takes as inputs the feature maps from the previous convolutional layers in the same block. For our experiments, we used DenseNet121, which is the shallowest variation as described in \cite{Gao_Huang}.

\textbf{SqueezeNet \cite{Forrest_Iandola}:} It reduces the number of parameters while maintaining competitive performance, this is achieved by designing so called Fire modules which contain two sub-modules, one squeezes information through a bottleneck and another one expands the amount of information. According to \cite{Forrest_Iandola}, SqueezeNet has similar performance to AlexNet on ImageNet but with overall 50x less parameters.

\textbf{MiniXception:} This network is a modification to the Xception network \cite{Francois_Chollet} since it reduces the amount of computation needed (e.g. for facial emotion recognition \cite{arriaga2017real}). Xception is a model that combines ideas from MobileNets by using depthwise separbale convolutions.

\clearpage
\section{Detailed RotNet Transfer Learning Results}
\label{sec:detailed_rotnet_transfer_learning_results}

In this section we show additional detailed transfer learning results with RotNet, presented in Table \ref{tab:rotnet_tl_accs}.

\begin{table*}[h]
\centering
\scriptsize
\begin{tabular}{llllll}
\toprule
\multicolumn{6}{l}{\textbf{Test Set Accuracy}}                                                                                                                                                                                                                                                                                                                                                                                                                                                                                                                                                                                                                                                                                                                                                                                                                                                                                                                                                                                                                                 \\
\midrule
\multicolumn{1}{l}{\textbf{RotNet Baseline}}                                                                                   & \multicolumn{1}{l}{\textbf{Layer}}                                                                                                                                                            & \multicolumn{1}{l}{\textbf{10 Samples}}                                                                                                                                                & \multicolumn{1}{l}{\textbf{40 Samples}}                                                                                                                                               & \multicolumn{1}{l}{\textbf{110 Samples}}                                                                                                                                              & \textbf{200 Samples}                                                                                                                                              \\ \hline
\multicolumn{1}{l}{\begin{tabular}[l]{@{}l@{}}ResNet20 Self-Supervised\\ \\ \\ \\ \\ ResNet20 Supervised\end{tabular}}         & \multicolumn{1}{l}{\begin{tabular}[l]{@{}l@{}}flatten\\ activation 18\\ activation 17\\ \\ \\ flatten\\ activation 18\\ activation 17\end{tabular}}                                           & \multicolumn{1}{l}{\begin{tabular}[l]{@{}l@{}}68.22  $\pm$  1.83\%\\ 66.64  $\pm$  2.95\%\\ 66.19  $\pm$  2.14\%\\ \\ \\ 76.51  $\pm$  3.30\%\\ 70.14  $\pm$  4.40\%\\ 71.61  $\pm$  3.10\%\end{tabular}} & \multicolumn{1}{l}{\begin{tabular}[l]{@{}l@{}}84.40  $\pm$  1.74\%\\ 85.70  $\pm$  0.84\%\\ 85.28  $\pm$  1.31\%\\ \\ \\ 90.20 $\pm$  1.23\%\\ 88.06  $\pm$  1.40\%\\ 88.43  $\pm$ 1.44\%\end{tabular}} & \multicolumn{1}{l}{\begin{tabular}[l]{@{}l@{}}90.06  $\pm$  1.08\%\\ 93.48  $\pm$  0.67\%\\ 93.62  $\pm$  0.83\%\\ \\ \\ 94.62  $\pm$  0.75\%\\ 95.27  $\pm$  0.74\%\\ 95.25  $\pm$  0.87\%\end{tabular}} & \begin{tabular}[l]{@{}l@{}}92.07  $\pm$  0.58\%\\ 96.46  $\pm$  0.54\%\\ \textbf{96.62  $\pm$  0.56\%}\\ \\ \\ 96.17  $\pm$  0.46\%\\ \textbf{97.27 $\pm$  0.60\%}\\ 97.14  $\pm$  0.79\%\end{tabular} \\
\midrule
\multicolumn{1}{l}{\begin{tabular}[l]{@{}l@{}}MobileNet Self-Supervised\\ \\ \\ \\ \\ MobileNet Supervised\end{tabular}}       & \multicolumn{1}{l}{\begin{tabular}[l]{@{}l@{}}conv\_pw\_11\_relu\\ flatten\\ conv\_pw\_12\_relu\\ \\ \\ conv\_pw\_11\_relu\\ flatten\\ conv\_pw\_12\_relu\end{tabular}}                       & \multicolumn{1}{l}{\begin{tabular}[l]{@{}l@{}}58.21  $\pm$  1.65\%\\ 51.08  $\pm$  2.07\%\\ 54.97  $\pm$  2.03\%\\ \\ \\ 63.90  $\pm$  1.64\%\\ 60.04 $\pm$  3.15\%\\ 59.98 $\pm$ 1.89\%\end{tabular}}     & \multicolumn{1}{l}{\begin{tabular}[l]{@{}l@{}}74.66  $\pm$  1.13\%\\ 68.32 $\pm$  1.28\%\\ 70.63 $\pm$  1.23\%\\ \\ \\ 76.38 $\pm$  1.72\%\\ 72.10  $\pm$  1.10\%\\ 74.35  $\pm$  1.35\%\end{tabular}}  & \multicolumn{1}{l}{\begin{tabular}[l]{@{}l@{}}83.86  $\pm$  1.03\%\\ 76.68  $\pm$  0.99\%\\ 79.30 $\pm$  1.24\%\\ \\ \\ 84.09  $\pm$  1.27\%\\ 78.80 $\pm$ 0.86\%\\ 80.40 $\pm$  1.27\%\end{tabular}}   & \begin{tabular}[l]{@{}l@{}}88.14  $\pm$  0.76\%\\ 82.35  $\pm$  0.99\%\\ 83.31  $\pm$  0.77\%\\ \\ \\ 87.34  $\pm$  0.84\%\\ 81.64  $\pm$  1.16\%\\ 83.69  $\pm$  0.48\%\end{tabular}   \\
\midrule
\multicolumn{1}{l}{\begin{tabular}[l]{@{}l@{}}DenseNet121 Self-Supervised\\ \\ \\ \\ \\ DenseNet121Supervised\end{tabular}}    & \multicolumn{1}{l}{\begin{tabular}[l]{@{}l@{}}conv5\_block15\_0\_relu\\ conv5\_block16\_0\_relu\\ avg pool\\ \\ \\ conv5\_block15\_0\_relu\\ conv5\_block16\_0\_relu\\ avg pool\end{tabular}} & \multicolumn{1}{l}{\begin{tabular}[l]{@{}l@{}}67.95 $\pm$ 2.17\%\\ 68.85 $\pm$ 2.88\%\\ 64.51 $\pm$  3.17\%\\ \\ \\ 69.95  $\pm$  3.29\%\\ 68.97 $\pm$  3.34\%\\ 63.65 $\pm$  2.70\%\end{tabular}}       & \multicolumn{1}{l}{\begin{tabular}[l]{@{}l@{}}85.59  $\pm$  1.20\%\\ 86.89 $\pm$  1.13\%\\ 80.96 $\pm$  1.11\%\\ \\ \\ 86.69 $\pm$  0.94\%\\ 86.26 $\pm$  1.38\%\\ 81.82 $\pm$  1.39\%\end{tabular}}     & \multicolumn{1}{l}{\begin{tabular}[l]{@{}l@{}}92.10 $\pm$  0.90\%\\ 92.58 $\pm$  0.73\%\\ 88.26 $\pm$  1.05\%\\ \\ \\ 93.75 $\pm$  0.64\%\\ 93.56 $\pm$  0.79\%\\ 88.60 $\pm$  0.85\%\end{tabular}}      & \begin{tabular}[l]{@{}l@{}}95.02  $\pm$  0.43\%\\ 94.66  $\pm$  0.56\%\\ 90.17  $\pm$  0.69\%\\ \\ \\ 95.70 $\pm$  0.41\%\\ 95.61 $\pm$  0.55\%\\ 91.96 $\pm$  0.37\%\end{tabular}    \\ \hline
\multicolumn{1}{l}{\begin{tabular}[l]{@{}l@{}}SqueezeNet Self-Supervised\\ \\ \\ \\ \\ SqueezeNet Supervised\end{tabular}}     & \multicolumn{1}{l}{\begin{tabular}[l]{@{}l@{}}batch\_norm\_8\\ batch\_norm\_9\\ global\_average\_pooling2d\\ \\ \\ batch\_norm\_8\\ batch\_norm\_9\\ global\_average\_pooling2d\end{tabular}} & \multicolumn{1}{l}{\begin{tabular}[l]{@{}l@{}}62.54  $\pm$  2.40\%\\ 39.70 $\pm$  2.99\%\\ 23.92 $\pm$  2.63\%\\ \\ \\ 71.76 $\pm$  2.61\%\\ 56.07 $\pm$  2.52\%\\ 38.77 $\pm$  2.61\%\end{tabular}}      & \multicolumn{1}{l}{\begin{tabular}[l]{@{}l@{}}82.10 $\pm$ 0.88\%\\ 55.14 $\pm$ 1.62\%\\ 25.13 $\pm$  0.98\%\\ \\ \\ 87.44 $\pm$  1.40\%\\ 73.59 $\pm$  0.96\%\\ 45.34 $\pm$ 1.59\%\end{tabular}}         & \multicolumn{1}{l}{\begin{tabular}[l]{@{}l@{}}89.56  $\pm$  0.85\%\\ 61.64 $\pm$  1.53\%\\ 26.27 $\pm$  0.80\%\\ \\ \\ 93.67 $\pm$  0.92\%\\ 80.34 $\pm$ 1.53\%\\ 47.22 $\pm$ 1.23\%\end{tabular}}       & \begin{tabular}[l]{@{}l@{}}92.37 $\pm$ 0.58\%\\ 64.65 $\pm$ 1.33\%\\ 25.95 $\pm$  0.98\%\\ \\ \\ 96.31  $\pm$  0.40\%\\ 83.07 $\pm$  0.84\%\\ 48.20 $\pm$  1.15\%\end{tabular}        \\
\midrule
\multicolumn{1}{l}{\begin{tabular}[l]{@{}l@{}}Minixception Self-Supervised\\ \\ \\ \\ \\ Minixception Supervised\end{tabular}} & \multicolumn{1}{l}{\begin{tabular}[l]{@{}l@{}}add\_3\\ add\_2\\ conv2d\_6\\ \\ \\ add\_3\\ add\_2\\ conv2d\_6\end{tabular}}                                                                   & \multicolumn{1}{l}{\begin{tabular}[l]{@{}l@{}}68.63 $\pm$ 2.13\%\\ 66.71 $\pm$  2.28\%\\ 52.74  $\pm$  2.43\%\\ \\ \\ 71.33  $\pm$  3.39\%\\ 70.34  $\pm$  2.23\%\\ 64.24  $\pm$  3.36\%\end{tabular}}   & \multicolumn{1}{l}{\begin{tabular}[l]{@{}l@{}}82.94 $\pm$ 1.41\%\\ 85.78 $\pm$ 1.21\%\\ 65.75 $\pm$ 1.48\%\\ \\ \\ 88.38 $\pm$ 0.97\%\\ 87.98 $\pm$ 0.96\%\\ 79.87 $\pm$ 1.20\%\end{tabular}}           & \multicolumn{1}{l}{\begin{tabular}[l]{@{}l@{}}90.12 $\pm$ 0.73\%\\ 92.80 $\pm$ 1.01\%\\ 70.91 $\pm$ 1.01\%\\ \\ \\ 94.34 $\pm$ 0.83\%\\ 94.63 $\pm$ 0.77\%\\ 86.68 $\pm$ 1.17\%\end{tabular}}             & \begin{tabular}[l]{@{}l@{}}92.29  $\pm$  0.70\%\\ 95.30 $\pm$  0.51\%\\ 73.73 $\pm$  0.95\%\\ \\ \\ 96.86 $\pm$ 0.46\%\\ 96.91 $\pm$ 0.68\%\\ 89.39 $\pm$ 0.56\%\end{tabular}     \\
\midrule
\multicolumn{1}{l}{Linear SVM}                                                                                                 & \multicolumn{1}{l}{NA}                                                                                                                                                                        & \multicolumn{1}{l}{63.55  $\pm$  3.66\%}                                                                                                                                                   & \multicolumn{1}{l}{85.39 $\pm$ 1.07\%}                                                                                                                                                   & \multicolumn{1}{l}{92.51 $\pm$ 0.79\%}                                                                                                                                                   & 95.67 $\pm$ 0.90\%                                                                                                                                                   \\
\bottomrule
\end{tabular}
\caption{RotNet transfer learning classification accuracies on the Turntable dataset.}
\label{tab:rotnet_tl_accs}
\end{table*}

\FloatBarrier
\clearpage
\section{Denoising Autoencoder Model Selection}
\label{sec:dae_model_selection}

In this section we describe the neural network architecture that we used as a Denoising Autoencoder.\\

We selected the following encoder-decoder CNN architecture for the  DAE, which based on the one reported in \cite{Matias_Valdenegro_IV}: The encoder architecture consists of Conv2D(32, $3 \times 3$) - MaxPool($2 \times 2$) - Conv2D(16, $3 \times 3$) - MaxPool($2 \times 2$) - Conv2D(8, $3 \times 3$) - MaxPool($2 \times 2$) - Flatten() - Fully Connected(c). The decoder architecture is composed by Fully Connected($c \times n_w \times n_h$) - Reshape() - Conv2D(32, $3 \times 3$) - UpSample($2 \times 2$) - Conv2D(16, $3 \times 3$) - UpSample($2 \times 2$) - Conv2D(8, $3 \times 3$) - UpSample($2 \times 2$) - Conv2D(1, $3 \times 3$).

\section{Denoising Autencoder Transfer Learning Results}
\label{sec:detailed_dae_transfer_learning_results}

In this section we show additional detailed transfer learning results with a Denoising Autoencoder, presented in Table \ref{tab:dae_tl_accs}.

\begin{table*}[h]
\centering
\scriptsize
\begin{tabular}{lllllll}
\toprule
\multicolumn{7}{l}{\textbf{Test Set Accuracy}}                                                                                                                                                                                                                                                                                                                                                                                                                                                                                                                                                                                                                                                                                                                                                                                                  \\
\midrule
\multicolumn{1}{l}{\textbf{Model}}                & \multicolumn{1}{l}{\textbf{Code Size}} & \multicolumn{1}{l}{\textbf{Gaussian Noise ($\sigma$)}}                                                 & \multicolumn{1}{l}{\textbf{10 Samples}}                                                                                                                       & \multicolumn{1}{l}{\textbf{40 Samples}}                                                                                                                       & \multicolumn{1}{l}{\textbf{110 Samples}}                                                                                                                       & \textbf{200 Samples}                                                                                                                     \\ 
\midrule
\multicolumn{1}{l}{\multirow{3}{*}{Denoising AE}} & \multicolumn{1}{l}{32}                 & \multicolumn{1}{l}{\begin{tabular}[c]{@{}c@{}}0.100\\ 0.125\\ 0.150\\ 0.175\\ 0.200\end{tabular}} & \multicolumn{1}{l}{\begin{tabular}[c]{@{}c@{}}64.40  $\pm$  2.98\%\\ 62.29  $\pm$  1.55\%\\ 60.89  $\pm$  2.62\%\\ 63.03  $\pm$  3.32\%\\ 61.59  $\pm$  2.35\%\end{tabular}} & \multicolumn{1}{l}{\begin{tabular}[c]{@{}c@{}}80.01  $\pm$  1.72\%\\ 75.42  $\pm$  1.02\%\\ 68.91  $\pm$  1.82\%\\ 77.28  $\pm$  1.11\%\\ 74.21  $\pm$  2.13\%\end{tabular}} & \multicolumn{1}{l}{\begin{tabular}[c]{@{}c@{}}82.68  $\pm$  1.20\%\\ 78.37  $\pm$  0.84\%\\  71.73  $\pm$  0.75\%\\ 81.20  $\pm$  1.21\%\\ 79.16  $\pm$  1.17\%\end{tabular}} & \begin{tabular}[c]{@{}c@{}}84.49  $\pm$  0.64\%\\ 79.48  $\pm$  1.00\%\\ 72.55 $\pm$  1.18\%\\ 82.62  $\pm$  0.85\%\\ 80.40  $\pm$  1.15\%\end{tabular} \\
\cmidrule{2-7}
\multicolumn{1}{l}{}                              & \multicolumn{1}{l}{64}                 & \multicolumn{1}{l}{\begin{tabular}[c]{@{}c@{}}0.100\\ 0.125\\ 0.150\\ 0.175\\ 0.200\end{tabular}} & \multicolumn{1}{l}{\begin{tabular}[c]{@{}c@{}}64.97  $\pm$  1.80\%\\ 61.81  $\pm$  2.59\%\\ 67.90  $\pm$  2.80\%\\ 63.93  $\pm$  3.15\%\\ 62.18  $\pm$  1.92\%\end{tabular}} & \multicolumn{1}{l}{\begin{tabular}[c]{@{}c@{}}83.34  $\pm$  1.47\%\\ 78.06  $\pm$  1.24\%\\ 81.44  $\pm$  1.72\%\\ 79.70  $\pm$  1.95\%\\ 79.84 + - 1.69\%\end{tabular}} & \multicolumn{1}{l}{\begin{tabular}[c]{@{}c@{}}88.06  $\pm$  0.81\%\\ 82.71  $\pm$  1.29\%\\ 86.79  $\pm$  0.91\%\\ 85.84  $\pm$  0.80\%\\ 85.05  $\pm$  1.23\%\end{tabular}}  & \begin{tabular}[c]{@{}c@{}}89.65  $\pm$  0.71\%\\ 85.11  $\pm$  1.0\%\\ 88.40  $\pm$  0.66\%\\ 87.45  $\pm$  0.77\%\\ 86.69  $\pm$  0.63\%\end{tabular}   \\
\cmidrule{2-7}
\multicolumn{1}{l}{}                              & \multicolumn{1}{l}{128}                & \multicolumn{1}{l}{\begin{tabular}[c]{@{}c@{}}0.100\\ 0.125\\ 0.150\\ 0.175\\ 0.200\end{tabular}} & \multicolumn{1}{l}{\begin{tabular}[c]{@{}c@{}}68.31  $\pm$  1.60\%\\ 65.05  $\pm$  2.98\%\\ 69.39  $\pm$  2.24\%\\ 69.81  $\pm$  2.31\%\\ 66.88  $\pm$  2.98\%\end{tabular}}   & \multicolumn{1}{l}{\begin{tabular}[c]{@{}c@{}}85.28  $\pm$  1.32\%\\ 83.42  $\pm$  1.61\%\\ 85.38  $\pm$  1.60\%\\ 86.82  $\pm$  0.97\%\\ 84.51  $\pm$  1.52\%\end{tabular}}  & \multicolumn{1}{l}{\begin{tabular}[c]{@{}c@{}}91.14  $\pm$  1.09\%\\ 88.94  $\pm$  1.22\%\\ 92.40  $\pm$  0.87\%\\ 92.37  $\pm$  0.78\%\\ 91.48 $\pm$  0.60\%\end{tabular}}  & \begin{tabular}[c]{@{}c@{}}94.31  $\pm$  0.63\%\\ 90.07  $\pm$  0.90\%\\ \textbf{94.49  $\pm$  0.62\%}\\ 93.96  $\pm$  0.73\%\\ 93.73  $\pm$  0.94\%\end{tabular} \\
\midrule
\multicolumn{1}{l}{\multirow{3}{*}{AE}}           & \multicolumn{1}{l}{32}                 & \multicolumn{1}{l}{-}                                                                             & \multicolumn{1}{l}{67.53  $\pm$  2.03\%}                                                                                                                           & \multicolumn{1}{l}{81.56  $\pm$  1.55\%}                                                                                                                         & \multicolumn{1}{l}{86.54 $\pm$ 1.13\%}                                                                                                                             & 88.31  $\pm$  0.83\%                                                                                                                         \\
\cmidrule{2-7}
\multicolumn{1}{l}{}                              & \multicolumn{1}{l}{64}                 & \multicolumn{1}{l}{-}                                                                             & \multicolumn{1}{l}{67.10  $\pm$  3.01\%}                                                                                                                           & \multicolumn{1}{l}{83.84 $\pm$  1.54\%}                                                                                                                         & \multicolumn{1}{l}{89.16  $\pm$  0.92\%}                                                                                                                          & 91.39  $\pm$  0.53\%                                                                                                                        \\ \cmidrule{2-7}
\multicolumn{1}{l}{}                              & \multicolumn{1}{l}{128}                & \multicolumn{1}{l}{-}                                                                             & \multicolumn{1}{l}{67.86  $\pm$  1.53\%}                                                                                                                            & \multicolumn{1}{l}{83.25  $\pm$  1.59\%}                                                                                                                         & \multicolumn{1}{l}{90.54  $\pm$  1.30\%}                                                                                                                          & \textbf{92.21  $\pm$  0.59\%}                                                                                                                        \\
\bottomrule
\end{tabular}
\caption{Denoising Autoencoder transfer learning classification accuracies on the Turntable dataset.}
\label{tab:dae_tl_accs}
\end{table*}

\FloatBarrier
\clearpage
\section{Jigsaw Model Selection}
\label{sec:jigsaw_model_selection}

In this section we describe the neural network architecture that we used for Jigsaw self-supervised learning.\\

Extensive experiments were carried out to optimize the CNN feature extractor design that works as a baseline for Jigsaw. The main parameters that we varied were the number of layers, number of filters and downsampling in the feature extractor. The original Jigsaw architecture \cite{Noroozi_I} uses a set of $\{64, 128, 256, 386\}$ 2DConvs. In our case, we have reduced considerably the number these filters (and hence number of parameters) to a set of  $\{32, 16, 8\}$ 2DConvs, this is because our Jigsaw model is trained on the smaller Watertank dataset and such large models might lead easily to overfitting.

We used a Time Distributed Layer\footnote{\url{https://keras.io/api/layers/recurrent_layers/time_distributed/}} (TDL) from Keras in order to feed image patches simultaneously through the sequential CNN feature extractor and a final decision network (classification layer). 
The feature extractor is composed of the following layers: Conv2D(32, $3 \times 3$) - BatchNorm() - MaxPool($2 \times 2$) - Dropout() - Conv2D(16, $3 \times 3$) - BatchNorm() - MaxPool($2 \times 2$) - Dropout() - Conv2D(8, $3 \times 3$) -BatchNorm() - MaxPool($2 \times 2$)- Dropout() - Flatten(). The TDL takes this sequential model and flattens the output predictions from every image tile to then process them through the decision network. The decision network is composed by: TimeDistributedLayer(9, None) - Flatten() - FullyConnected() - BatchNorm() - FullyConnected() - BatchNorm() - Dropout() - FullyConnected().

\section{Jigsaw Transfer Learning Results}
\label{sec:detailed_jigsaw_transfer_learning_results}

In this section we show additional detailed transfer learning results with Jigsaw, presented in Table \ref{tab:jigsaw_tl_accs}.

\begin{table*}[h]
\centering
\scriptsize
\begin{tabular}{llllll}
\toprule
\multicolumn{6}{l}{\textbf{Test Set Accuracy}}                                                                                                                                                                                                                          \\
\midrule
\multicolumn{1}{l}{\textbf{Permutations}}       & \multicolumn{1}{l}{\textbf{Layer}} & \multicolumn{1}{l}{\textbf{10 Samples}} & \multicolumn{1}{l}{\textbf{40 Samples}} & \multicolumn{1}{l}{\textbf{110 Samples}} & \multicolumn{1}{l}{\textbf{200 Samples}}     \\
\midrule
\multicolumn{1}{l}{\multirow{3}{*}{5}}          & \multicolumn{1}{l}{dropout\_0}     & \multicolumn{1}{l}{60.61 $\pm$ 1.68\%}      & \multicolumn{1}{l}{83.93 $\pm$ 1.25\%}      & \multicolumn{1}{l}{91.59 $\pm$ 0.63\%}      & \multicolumn{1}{l}{95.19 $\pm$ 1.09\%}          \\ \cmidrule{2-6}
\multicolumn{1}{l}{}                            & \multicolumn{1}{l}{dropout\_1}     & \multicolumn{1}{l}{66.41 $\pm$ 2.67\%}     & \multicolumn{1}{l}{86.17 $\pm$ 1.06\%}     & \multicolumn{1}{l}{94.51 $\pm$ 0.95\%}      & \multicolumn{1}{l}{96.74 $\pm$ 0.66\%}          \\ \cmidrule{2-6}
\multicolumn{1}{l}{}                            & \multicolumn{1}{l}{dropout\_2}     & \multicolumn{1}{l}{70.14 $\pm$ 2.36\%}      & \multicolumn{1}{l}{86.76 $\pm$ 2.10\%}      & \multicolumn{1}{l}{94.35 $\pm$ 1.16\%}      & \multicolumn{1}{l}{96.24 $\pm$ 0.47\%}          \\
\midrule
\multicolumn{1}{l}{\multirow{3}{*}{10}}         & \multicolumn{1}{l}{dropout\_0}     & \multicolumn{1}{l}{64.248 $\pm$ 2.22\%}     & \multicolumn{1}{l}{84.71 $\pm$ 1.79\%}     & \multicolumn{1}{l}{92.65 $\pm$ 1.44\%}      & \multicolumn{1}{l}{95.62 $\pm$ 0.74\%}          \\ \cmidrule{2-6}
\multicolumn{1}{l}{}                            & \multicolumn{1}{l}{dropout\_1}     & \multicolumn{1}{l}{64.62 $\pm$ 1.75\%}      & \multicolumn{1}{l}{86.29 $\pm$ 1.85\%}     & \multicolumn{1}{l}{94.76 $\pm$ 0.74\%}       & \multicolumn{1}{l}{\textbf{96.83 $\pm$ 0.47 \%}}   \\ \cmidrule{2-6}
\multicolumn{1}{l}{}                            & \multicolumn{1}{l}{dropout\_2}     & \multicolumn{1}{l}{66.20 $\pm$ 1.51\%}      & \multicolumn{1}{l}{88.55 $\pm$ 0.78\%}     & \multicolumn{1}{l}{93.95 $\pm$ 1.41\%}      & \multicolumn{1}{l}{96.77 $\pm$ 0.41\%}          \\
\midrule
\multicolumn{1}{l}{\multirow{3}{*}{15}}         & \multicolumn{1}{l}{dropout\_0}     & \multicolumn{1}{l}{64.0 $\pm$ 3.88\%}       & \multicolumn{1}{l}{84.40 $\pm$ 0.98\%}     & \multicolumn{1}{l}{91.84 $\pm$ 1.13\%}      & \multicolumn{1}{l}{95.16 $\pm$ 0.50\%}          \\ \cmidrule{2-6}
\multicolumn{1}{l}{}                            & \multicolumn{1}{l}{dropout\_1}     & \multicolumn{1}{l}{67.10 $\pm$ 1.68\%}      & \multicolumn{1}{l}{86.73 $\pm$ 1.66\%}     & \multicolumn{1}{l}{94.51 $\pm$ 1.07\%}      & \multicolumn{1}{l}{96.65 $\pm$ 0.46\%}            \\ \cmidrule{2-6}
\multicolumn{1}{l}{}                            & \multicolumn{1}{l}{dropout\_2}     & \multicolumn{1}{l}{68.18 $\pm$ 4.74\%}     & \multicolumn{1}{l}{85.73 $\pm$ 2.09\%}     & \multicolumn{1}{l}{93.79 $\pm$ 0.68\%}      & \multicolumn{1}{l}{95.93 $\pm$ 0.58\%}          \\
\midrule
\multicolumn{1}{l}{\multirow{3}{*}{20}}         & \multicolumn{1}{l}{dropout\_0}     & \multicolumn{1}{l}{60.71 $\pm$ 2.55\%}     & \multicolumn{1}{l}{83.59  $\pm$ 0.73\%}     & \multicolumn{1}{l}{91.84 $\pm$ 0.77\%}       & \multicolumn{1}{l}{95.19 $\pm$ 0.39\%}          \\ \cmidrule{2-6}
\multicolumn{1}{l}{}                            & \multicolumn{1}{l}{dropout\_1}     & \multicolumn{1}{l}{63.16 $\pm$ 1.89\%}      & \multicolumn{1}{l}{85.36 $\pm$ 1.51\%}     & \multicolumn{1}{l}{93.55 $\pm$ 0.63\%}       & \multicolumn{1}{l}{95.62 $\pm$ 0.87\%}          \\ \cmidrule{2-6}
\multicolumn{1}{l}{}                            & \multicolumn{1}{l}{dropout\_2}     & \multicolumn{1}{l}{67.25 $\pm$ 2.57\%}     & \multicolumn{1}{l}{87.16 $\pm$ 1.34\%}     & \multicolumn{1}{l}{93.36 $\pm$ 0.72\%}      & \multicolumn{1}{l}{95.59 $\pm$ 0.52\%}          \\
\midrule
\multicolumn{1}{l}{\multirow{3}{*}{Supervised}} & \multicolumn{1}{l}{dropout\_0}     & \multicolumn{1}{l}{65.45 $\pm$ 2.96\%}     & \multicolumn{1}{l}{87.59 $\pm$ 1.51\%}     & \multicolumn{1}{l}{93.82 $\pm$ 1.05\%}      & \multicolumn{1}{l}{96.58 $\pm$ 0.59\%}          \\ \cmidrule{2-6}
\multicolumn{1}{l}{}                            & \multicolumn{1}{l}{dropout\_1}     & \multicolumn{1}{l}{69.64 $\pm$ 3.99\%}       & \multicolumn{1}{l}{89.33 $\pm$ 1.09\%}     & \multicolumn{1}{l}{94.91 $\pm$ 0.85\%}      & \multicolumn{1}{l}{\textbf{97.08 $\pm$ 0.22\%}} \\ \cmidrule{2-6}
\multicolumn{1}{l}{}                            & \multicolumn{1}{l}{dropout\_2}     & \multicolumn{1}{l}{66.94 $\pm$ 1.26\%}     & \multicolumn{1}{l}{87.25 $\pm$ 1.46\%}     & \multicolumn{1}{l}{93.89 $\pm$ 1.51\%}      & \multicolumn{1}{l}{96.09 $\pm$ 0.51\%}          \\
\bottomrule
\end{tabular}
\caption{Jigsaw transfer learning classification accuracies on the Turntable dataset.}
\label{tab:jigsaw_tl_accs}
\end{table*}

\FloatBarrier
\clearpage
\section{Comparison of Best Performing Transfer Learning Models}
\label{sec:detailed_best_transfer_learning_results}

\begin{figure*}[h]
  \centering
  \begin{subfigure}{0.4\linewidth}
  	\includegraphics[scale=0.14]{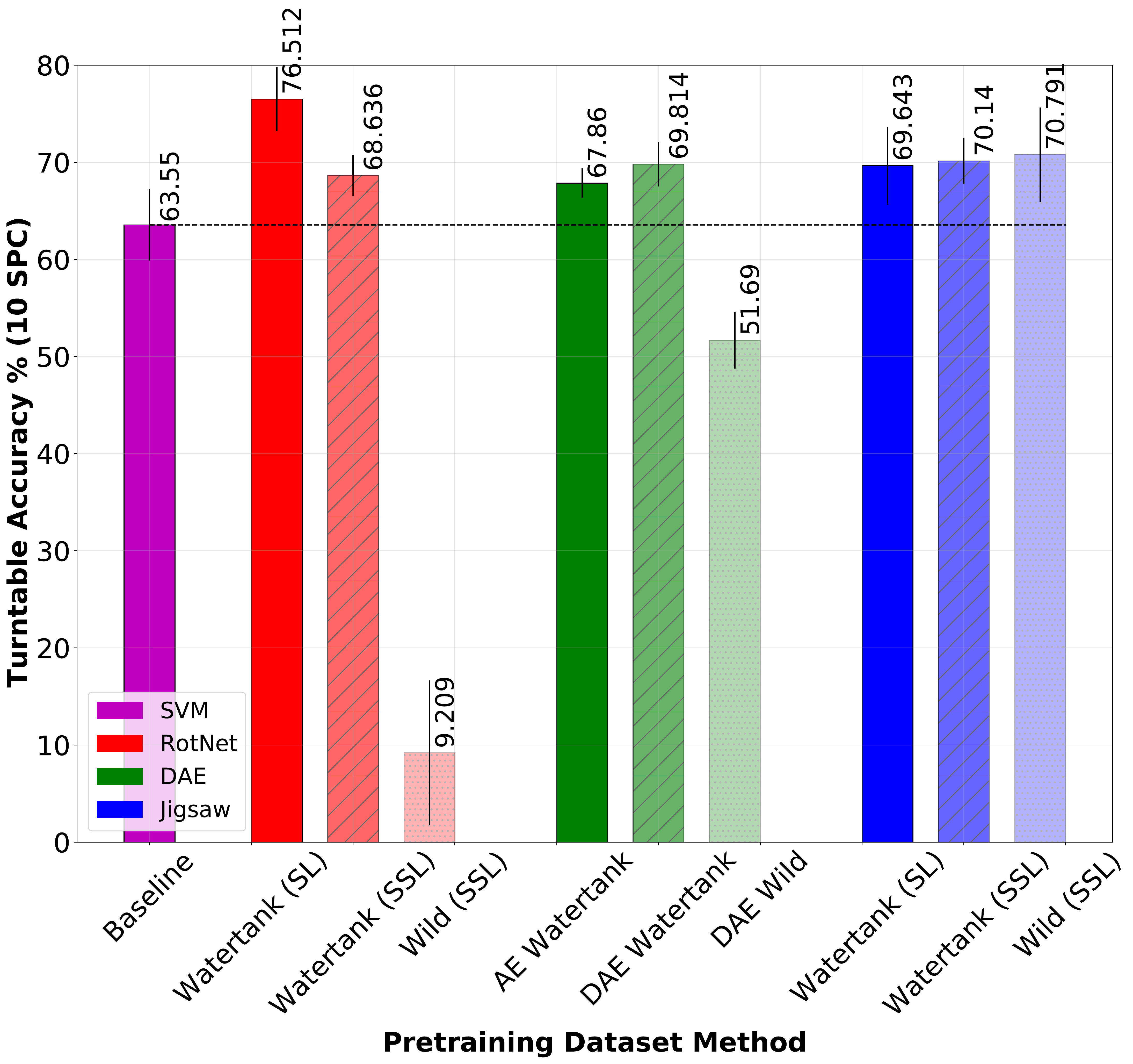}
    \caption{10 spc.}
    \label{fig:10_spc}
  \end{subfigure}
  \begin{subfigure}{0.4\linewidth}
    \includegraphics[scale=0.14]{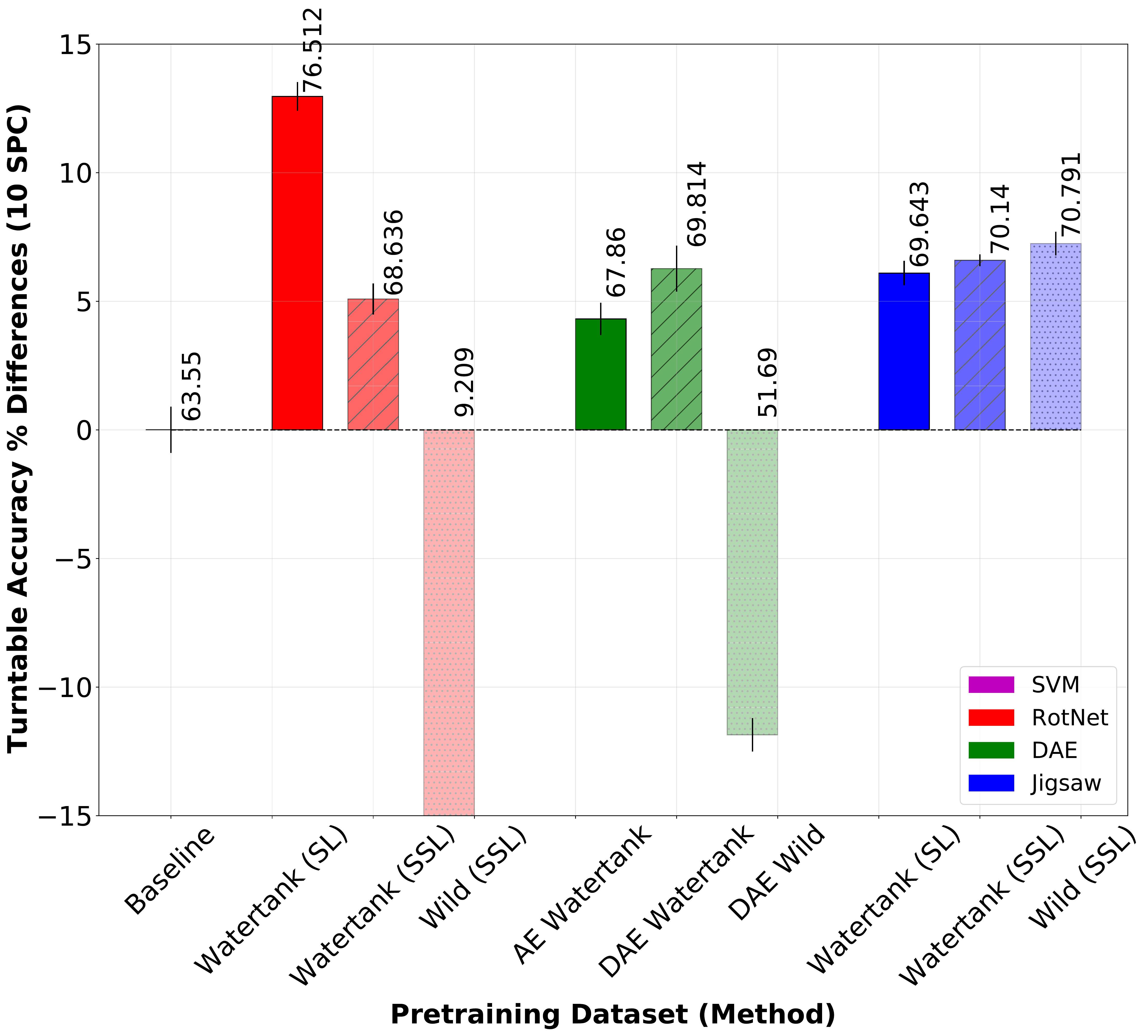}
    \caption{10 spc with baseline substracted.}
    \label{fig:10_spc_baseline}
  \end{subfigure}

  \bigskip %

  \centering
  \begin{subfigure}{0.4\linewidth}
  	\includegraphics[scale=0.14]{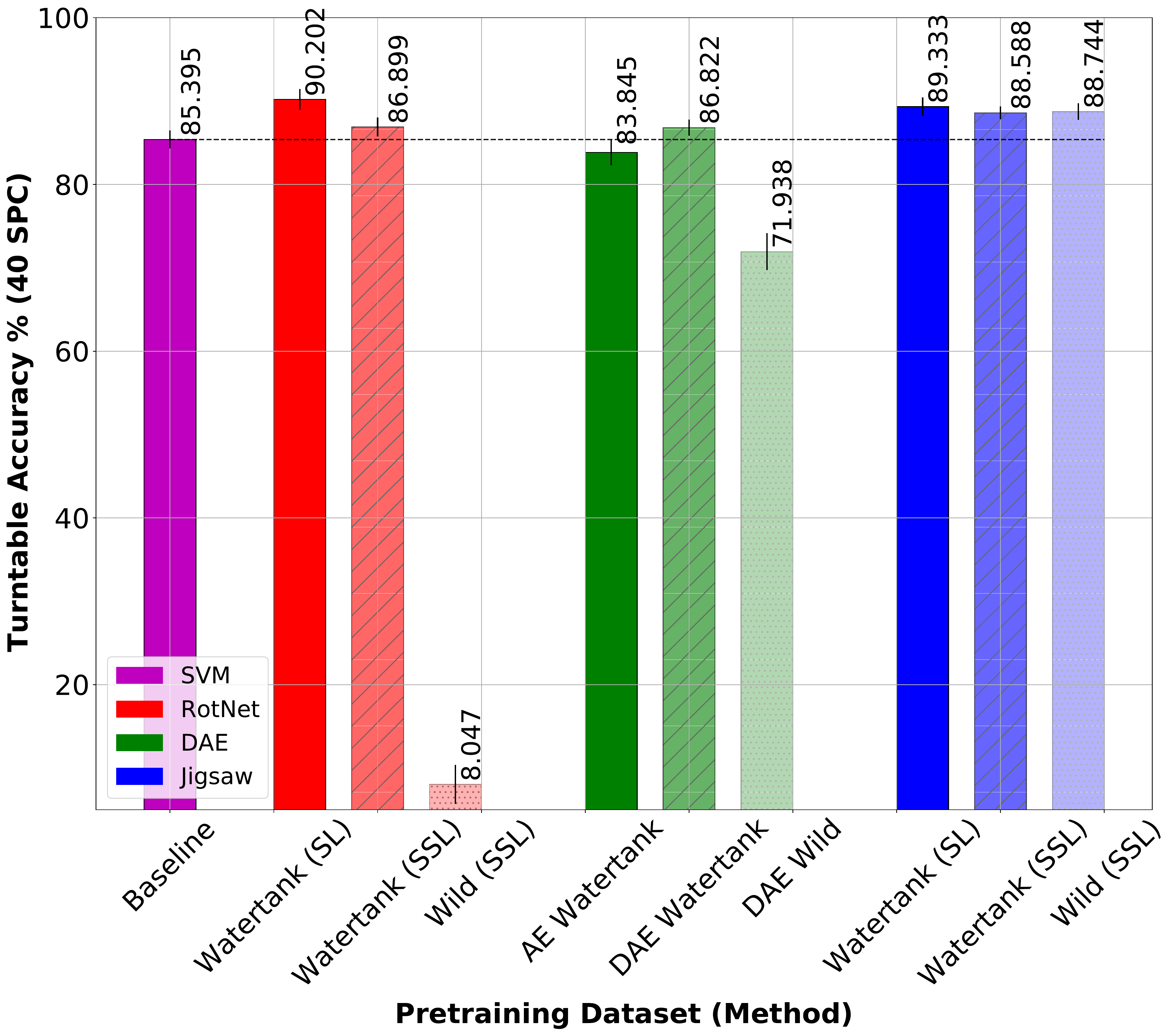}
    \caption{40 spc}
    \label{fig:40_spc}
  \end{subfigure}
  \begin{subfigure}{0.4\linewidth}
    \includegraphics[scale=0.14]{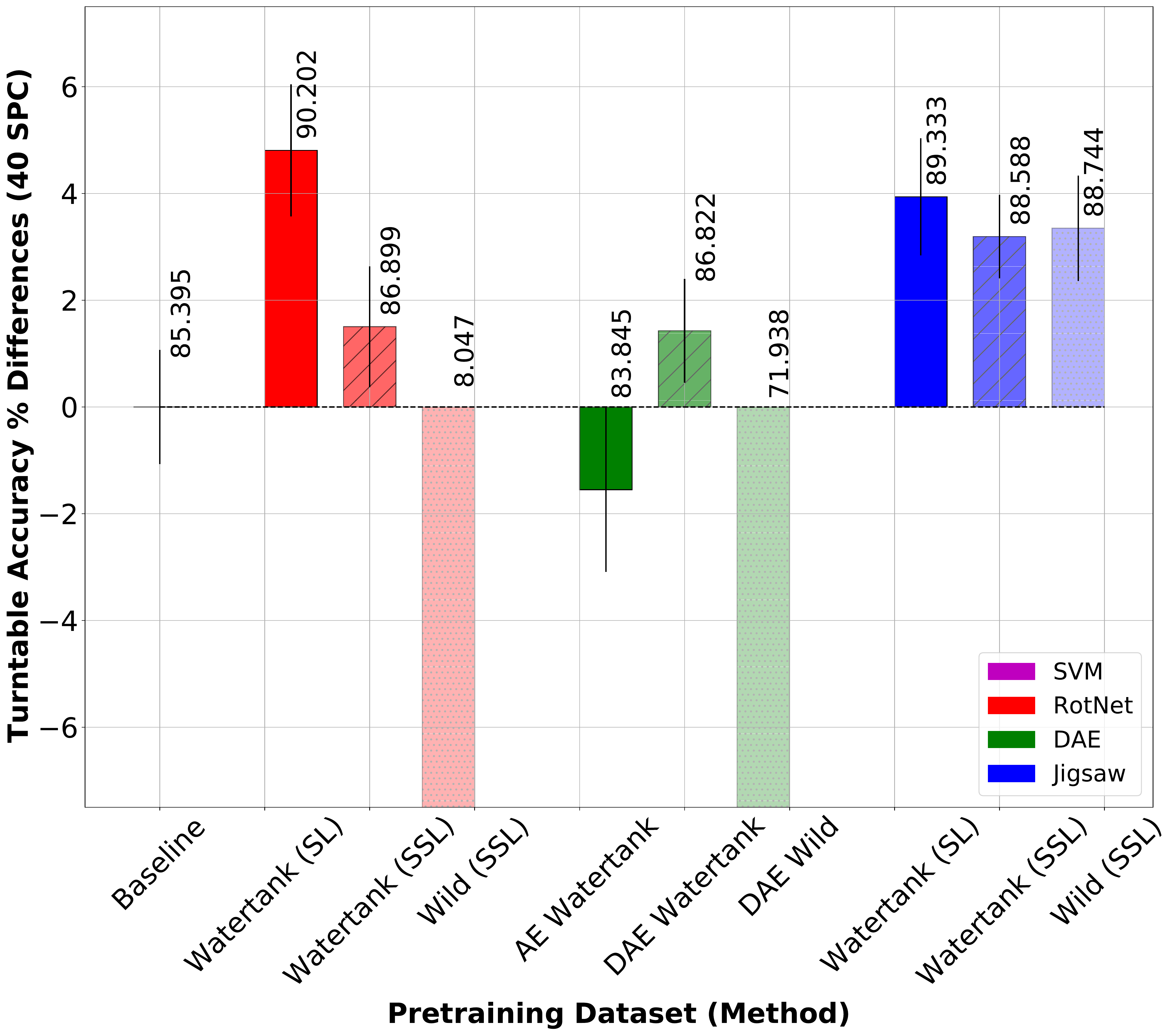}
    \caption{40 spc with baseline substracted.}
    \label{fig:40_spc_baseline}
  \end{subfigure}

    \bigskip %

  \centering
  \begin{subfigure}{0.4\linewidth}
  	\includegraphics[scale=0.14]{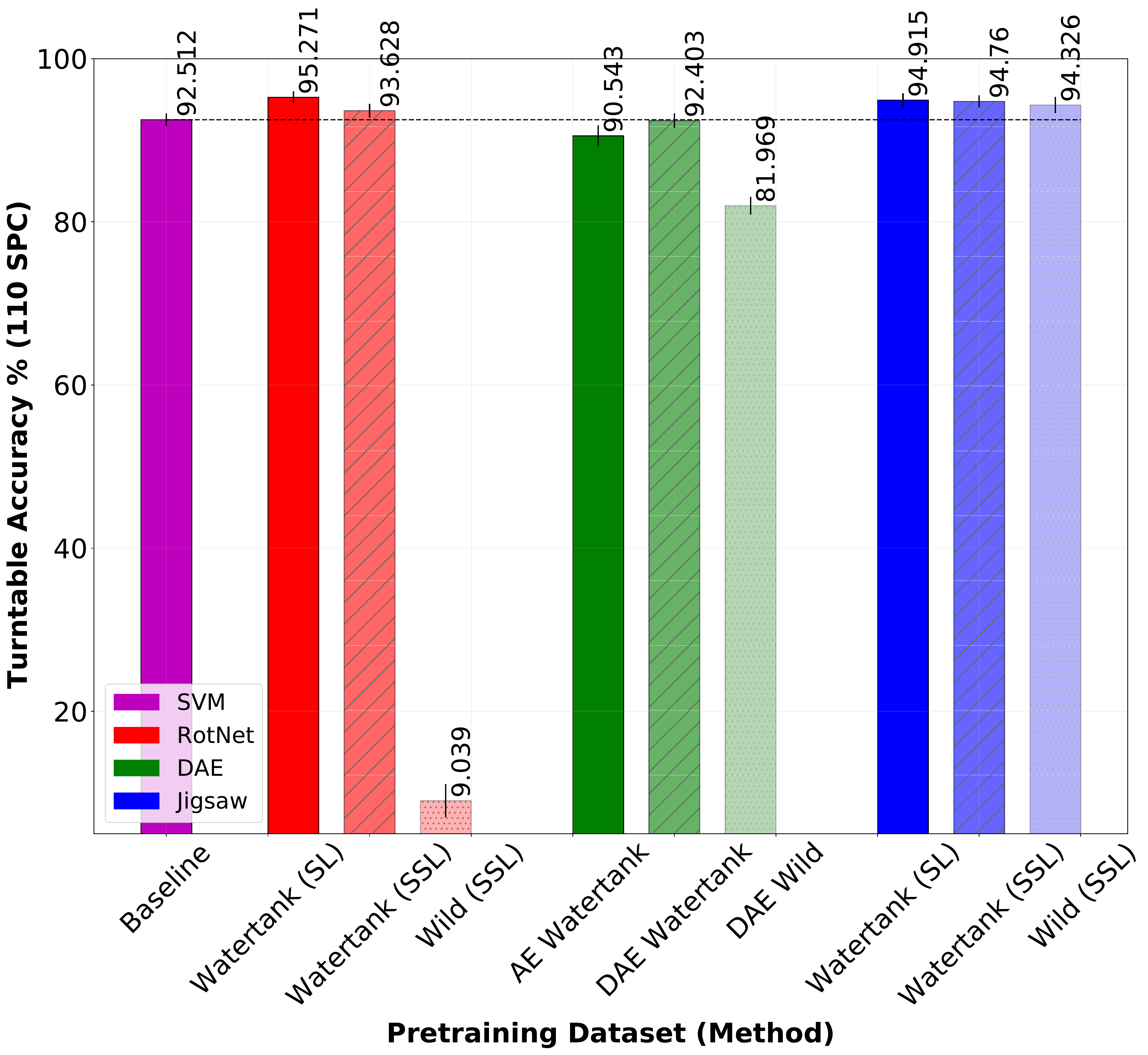}
    \caption{110 spc}
    \label{fig:110_spc}
  \end{subfigure}
  \begin{subfigure}{0.4\linewidth}
    \includegraphics[scale=0.14]{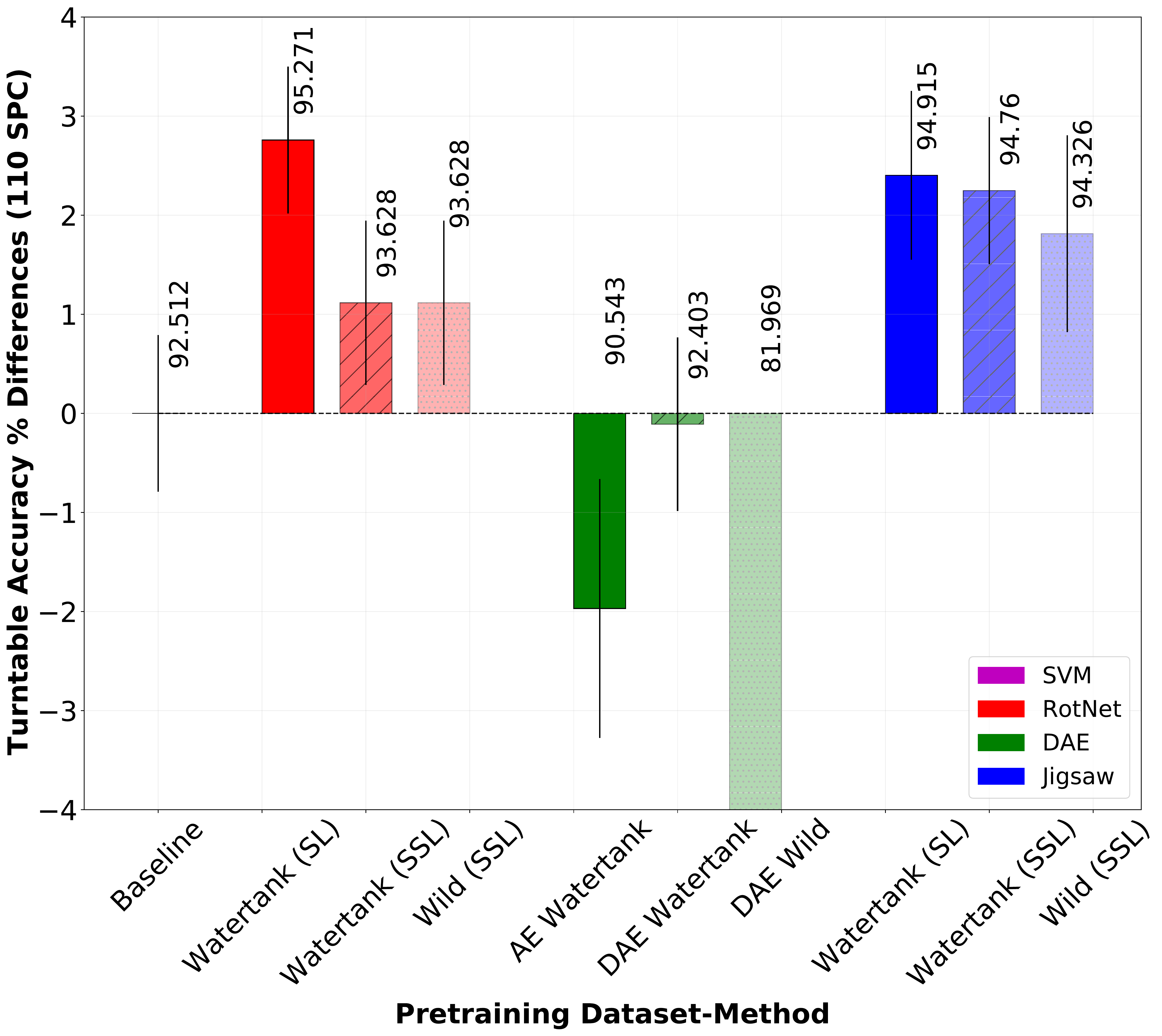}
    \caption{110 spc with baseline substracted.}
    \label{fig:110_spc_baseline}
  \end{subfigure}
  \caption{Comparison of the best performing SSL models pretrained on the Watertank sonar dataset (with supervision and self-supervision) and the Wild sonar dataset. 10, 40 and 110 spc cases.}
  \label{fig:best_performing_bars_TL}
\end{figure*}

In this section we show additional comparisons of the best transfer learning results, presented in Figure \ref{fig:best_performing_bars_TL} for selected values of samples per class (SPC).

\end{document}